%% file: main.tex
\documentclass[conference]{IEEEtran}

\usepackage{cite}
\usepackage{amsmath,amssymb,amsfonts}
\usepackage{algorithmic}
\usepackage{multicol}

\usepackage{graphicx}
\graphicspath{{figures/}}
\usepackage[labelformat=parens, labelsep=space]{subcaption}

\usepackage{textcomp}
\usepackage[table, xcdraw]{xcolor} 
\usepackage{diagbox}
\usepackage{multirow,booktabs,color,soul,threeparttable}
\definecolor{hl}{RGB}{255,255,0}
\sethlcolor{hl}

\usepackage{lettrine} 

\usepackage{flushend}

\usepackage{mathtools}

\DeclarePairedDelimiter\floor{\lfloor}{\rfloor}
\usepackage[linesnumbered,algoruled,boxed,lined]{algorithm2e}
\SetKwInOut{Parameter}{Parameters}
\newcommand*\mean[1]{\overline{#1}}

\renewcommand\IEEEkeywordsname{Keywords}

\def\BibTeX{{\rm B\kern-.05em{\sc i\kern-.025em b}\kern-.08em
    T\kern-.1667em\lower.7ex\hbox{E}\kern-.125emX}}

\makeatletter
\def\endthebibliography{%
  \def\@noitemerr{\@latex@warning{Empty `thebibliography' environment}}%
  \endlist
}

\IEEEoverridecommandlockouts
\def\footnoterule{\relax%
  \kern-5pt
  \hbox to \columnwidth{\hfill\vrule width 1\columnwidth height 0.4pt\hfill}
  \kern4.6pt}
\makeatother

\begin{document}
\title{A Decomposition-based Large-scale Multi-modal Multi-objective Optimization Algorithm
}
\author{\IEEEauthorblockN{Yiming Peng, Hisao Ishibuchi\IEEEauthorrefmark{2}}
\IEEEauthorblockA{Guangdong Provincial Key Laboratory of Brain-inspired Intelligent Computation,}
\IEEEauthorblockA{Department of Computer Science and Engineering, Southern University of Science and Technology, Shenzhen 518055, China}
\IEEEauthorblockA{11510035@mail.sustech.edu.cn, hisao@sustech.edu.cn}
\thanks{
This work was supported by Guangdong Provincial Key Laboratory (Grant No. 2020B121201001), the Program for Guangdong Introducing Innovative and Enterpreneurial Teams (Grant No. 2017ZT07X386), Shenzhen Science and Technology Program (Grant No. KQTD2016112514355531), the Program for University Key Laboratory of Guangdong Province (Grant No. 2017KSYS008).

\IEEEauthorrefmark{2}Corresponding author: Hisao Ishibuchi, hisao@sustech.edu.cn
}
}

\maketitle

\begin{abstract}
A multi-modal multi-objective optimization problem is a special kind of multi-objective optimization problem with multiple Pareto subsets. In this paper, we propose an efficient multi-modal multi-objective optimization algorithm based on the widely used MOEA/D algorithm. In our proposed algorithm, each weight vector has its own sub-population. With a clearing mechanism and a greedy removal strategy, our proposed algorithm can effectively preserve equivalent Pareto optimal solutions (i.e., different Pareto optimal solutions with same objective values). Experimental results show that our proposed algorithm can effectively preserve the diversity of solutions in the decision space when handling large-scale multi-modal multi-objective optimization problems.
\end{abstract}

\bigskip 


\input{sections/Introduction}
\input{sections/RelatedWorks}
\input{sections/Algorithm}
\input{sections/ExperimentalResults}
\input{sections/Conclusion}

\bibliographystyle{IEEEtran}
\bibliography{Reference.bib}
\end{document}

%% file: sections/Introduction.tex
\section{Introduction}

\label{sec: Introduction}
A multi-objective optimization problem (\textbf{MOP}) is an optimization problem which has multiple objective functions. Usually, these objective functions are conflicting and cannot be optimized simultaneously. For convenience, all objective functions should be converted into minimization functions. The following equation formulates an MOP without constraints:
\begin{equation}
    \min \boldsymbol{F}(\boldsymbol{x}) = (f_1(\boldsymbol{x}), \ldots, f_M(\boldsymbol{x}))^T,
    \label{eq: MOP formulation}
\end{equation}
where $\boldsymbol{x}$ is a $D$-dimensional decision vector, and $\boldsymbol{F}$ is a mapping from a $D$-dimensional domain $\Omega$ to an $M$-dimensional range $\mathbb{R}^M$.

In the past three decades, researchers have developed a variety of multi-objective evolutionary algorithms (\textbf{MOEAs}). For example, the well-known NSGA-II\cite{deb2002fast} and MOEA/D\cite{zhang2007moea} can efficiently solve various types of MOPs.

\begin{figure}
	\begin{subfigure}[b]{.24\textwidth}
		\includegraphics[width=\linewidth]{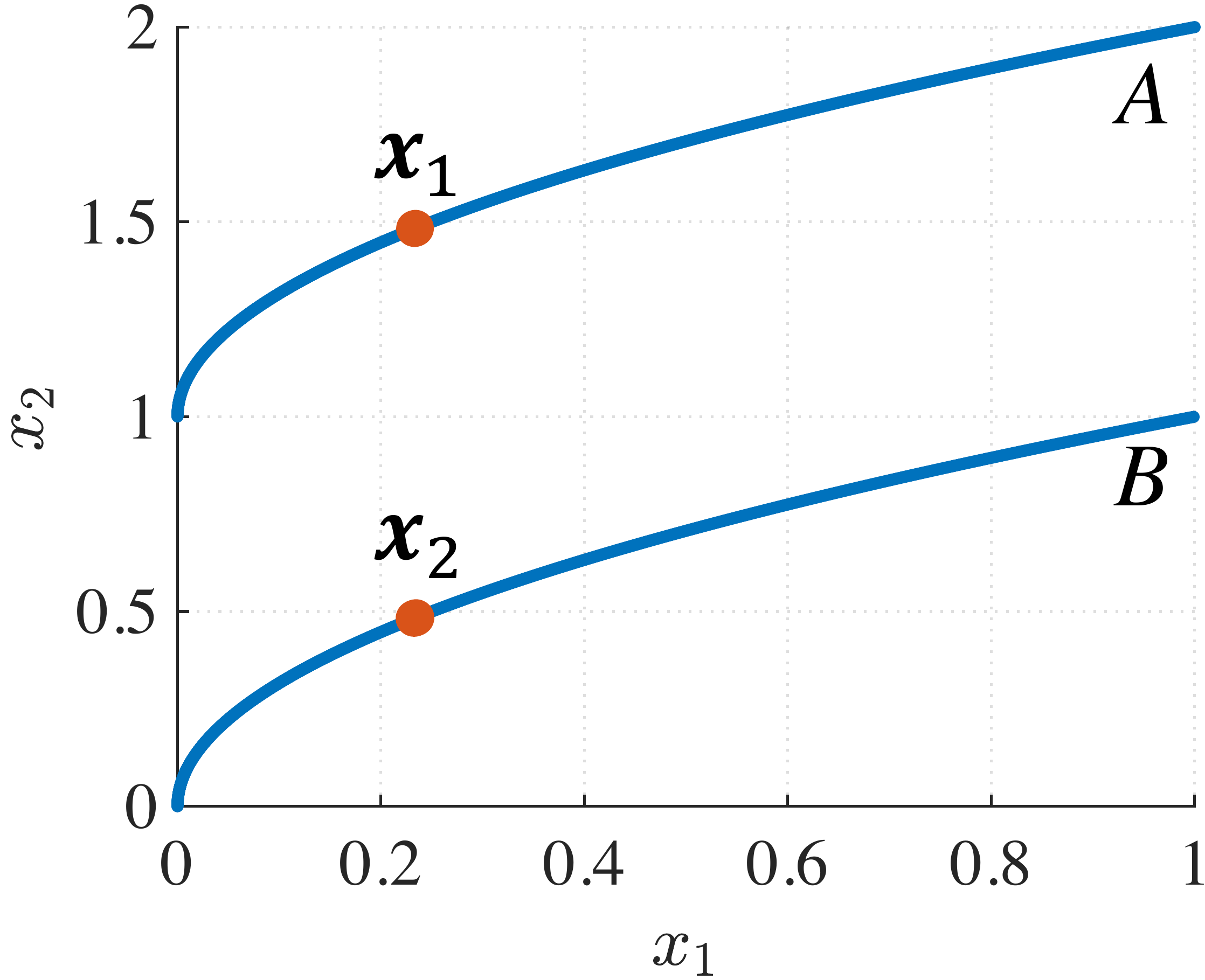}
		\caption{Decision space.}
	\end{subfigure}
	\begin{subfigure}[b]{.24\textwidth}
		\includegraphics[width=\linewidth]{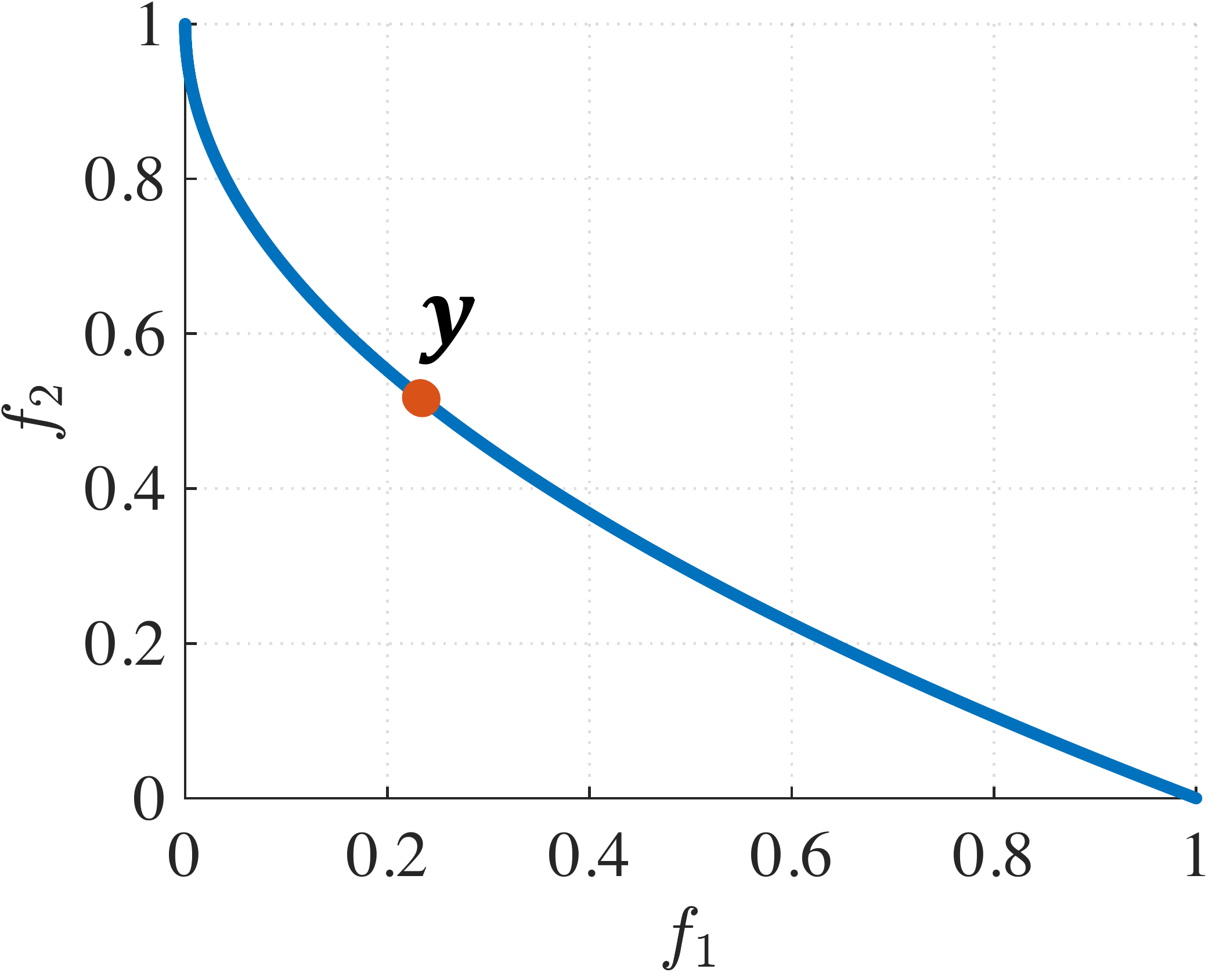}
		\caption{Objective space.}
	\end{subfigure}
	\caption{The two equivalent Pareto subsets ($A$ and $B$ in (a)) and the Pareto front (the blue curve in (b)) of the SUF3 test problem. Solution $\boldsymbol{y}$ has two inverse images $\boldsymbol{x}_1$ and $\boldsymbol{x}_2$ in the decision space.}
	\label{fig: SUF3}
\end{figure}

In this paper, we mainly focus on a special type of MOPs called multi-modal multi-objective optimization problems (\textbf{MMOPs}). In MMOPs, the function $\boldsymbol{F}$ in Eq. (\ref{eq: MOP formulation}) can be a many-to-one mapping from $\Omega$ to $\mathbb{R}^M$. That is, for a Pareto optimal solution in the objective space, there may exist multiple inverse images in the decision space. Formally, solutions $\boldsymbol{x}_1$ and $\boldsymbol{x}_2$ are equivalent iff $\boldsymbol{F}(\boldsymbol{x}_1) = \boldsymbol{F}(\boldsymbol{x}_2)$ and $\boldsymbol{x}_1 \neq \boldsymbol{x}_2$. As reported in \cite{moshaiov2016paradox}, local optimal solutions is not well-defined for MOPs. Therefore, we only consider global Pareto optimal solutions. Due to the existence of equivalent solutions, an MMOP may have multiple equivalent Pareto subsets, each of which is mapped to the whole Pareto front. For example, in Fig. \ref{fig: SUF3}, the SUF3 test problem \cite{liang2016multimodal} has two equivalent Pareto subsets $A$ and $B$ and each of them is mapped to the whole Pareto front in the objective space. When handling MMOPs, all equivalent Pareto subsets should be covered. A wide variety of real-world problems are MMOPs. For instance, the space mission design problems\cite{schutze2011computing} and the multi-objective knapsack problems\cite{jaszkiewicz2002performance} are MMOPs. Solving MMOPs is very useful since equivalent Pareto optimal solutions offer more alternatives for the decision maker\cite{deb2001multi}. 

As pointed out in the literature\cite{liang2016multimodal, tanabe2018decomposition}, MOEAs cannot efficiently solve MMOPs since they do not consider the diversity of solutions in the decision space. Therefore, several multi-modal multi-objective optimization evolutionary algorithms (\textbf{MMEAs}) are proposed such as DNEA\cite{liu2018double}, MO\_Ring\_PSO\_SCD\cite{yue2018multiobjective} and MOEA/D-AD\cite{tanabe2018decomposition}. As reported in \cite{peng2019multi}, these algorithms perform well on MMOPs with low-dimensional decision spaces. However, only a few of them can obtain a diverse solution set in the decision space when handling MMOPs with high-dimensional decision spaces (i.e., large-scale MMOPs). To alleviate this issue, we propose an efficient algorithm for large-scale MMOPs.

The rest of the paper is organized as follows. We briefly review related studies on decomposition-based MOEAs and MMEAs in Section \ref{sec: Related Works}. In Section \ref{sec: Proposed Algorithm}, the proposed MMEA is outlined. Then we compare the proposed MMEA with representative state-of-the-art algorithms on several MMOPs in Section \ref{sec: Experimental Results}. Finally, we summarize the paper and suggest some future research topics in Section \ref{sec: Concluding Remarks}.

%% file: sections/RelatedWorks.tex
\section{Related Work}
\label{sec: Related Works}
\subsection{Decomposition-based MOEAs}
\label{sec: Decomposition-based MOEAs}
In 2007, Zhang and Li proposed MOEA/D\cite{zhang2007moea}, the first decomposition-based MOEA. MOEA/D convert an MOP into a set of scalar optimization problems. For this reason, MOEA/D can maintain a strong search ability when handling MOPs with many objectives (i.e., many-objective optimization problems). In addition, MOEA/D is capable of obtaining a uniformly distributed solution set if the given weight vectors uniformly intersect with the Pareto front. Several efficient methods such as MOEA/D-AWA\cite{Qi2014adaptive} are proposed for the adaptive adjustment of weight vectors.

\begin{figure}
    \centering
    \includegraphics[width=.3\textwidth]{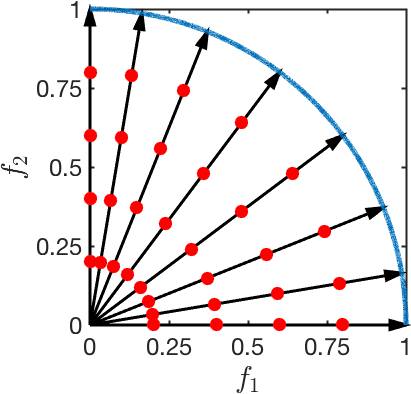}
    \caption{Explanation of MOEA/D based on the sub-population framework.}
    \label{fig: Explanation of MOEA/D with sub-population}
\end{figure}

\subsection{MOEA/D variants based on the sub-population framework}
\label{sec: MOEA/D variants based on sub-population framework}
Although MOEA/D shows promising performance on various types of MOPs, it is not suitable for solving MMOPs as reported in \cite{tanabe2018decomposition}. The main reason is that only one solution is assigned to each weight vector which guides a search direction to a Pareto optimal solution on the Pareto front. Thus, MOEA/D cannot preserve multiple equivalent solutions, i.e., is unable to solve MMOPs. A very straightforward remedy is to assign multiple solutions to each weight vector. In this way, each weight vector is assigned a sub-population instead of a single solution. In Fig. \ref{fig: Explanation of MOEA/D with sub-population}, each weight vector has a sub-population containing four solutions. Solutions in the same sub-population search toward equivalent Pareto optimal solutions. To our best knowledge, three MOEA/D variants use this sub-population framework: MOEA/D-AD\cite{tanabe2018decomposition}, MOEA/D-M2M\cite{Liu2014decomposition}, and an algorithm from \cite{Hu2018Incorporation}. MOEA/D-M2M is a standard MOEA while the other two algorithms are MMEAs. MOEA/D-M2M lacks a mechanism to preserve equivalent Pareto optimal solutions, which means that it is not suitable for solving MMOPs. Here we briefly introduce MOEA/D-AD and the algorithm from \cite{Hu2018Incorporation}.

In MOEA/D-AD, the sub-population size of each weight vector changes adaptively. The mechanism of MOEA/D-AD can be briefly described as follows. In each iteration,
\begin{enumerate}
    \item An offspring $\boldsymbol{y}$ is generated and assigned to the closest weight vector $\boldsymbol{w}_i$ in the objective space. The sub-population of $\boldsymbol{w}_i$ is denoted as $P_i$
    \item The closest $L$ solutions (denoted as $Q$) to $\boldsymbol{y}$ in the decision space are selected from the whole population. 
    \item The offspring $\boldsymbol{y}$ will be added to $P_i$ if one of the following two conditions is met:
    \begin{itemize}
        \item If $P_i\cap{}Q\neq\emptyset$, and at least one solution in $P_i\cap{}Q$ has the worse scalarizing function value than $\boldsymbol{y}$. Solutions worse than $\boldsymbol{y}$ will be removed from $P_i$
        \item If $P_i\cap{}Q=\emptyset$.
    \end{itemize}
\end{enumerate}
The niching structure in the decision space is maintained by the procedures described in step 3). The offspring $\boldsymbol{y}$ only competes with solutions in the current sub-population that are its neighbors in the decision space. In this way, the sub-population of a weight vector can keep multiple equivalent solutions. Since MOEA/D-AD uses an unbounded population, the size of the whole population may become very large after many iterations.

In the algorithm from \cite{Hu2018Incorporation}, each weight vector is assigned $k$ solutions. In this way, a population is separated into $k$ grids, each of which contains one solution from each weight vector. For each grid, the fitness of a solution $\boldsymbol{x}$ which is assigned to a weight vector $\boldsymbol{w}$ is evaluated based on the following equation: 
\begin{equation}
    f(\boldsymbol{x}) = w_1g(\boldsymbol{w}, \boldsymbol{x}) + w_2d_{min} + w_3d_{avg},
\end{equation}
where $g$ is an aggregation function, $d_{min}$ is the minimum Euclidean distance from $\boldsymbol{x}$ to other solutions assigned to $\boldsymbol{w}$, and $d_{avg}$ is the average Euclidean distance from $\boldsymbol{x}$ to solutions assigned to other weight vectors. These three functions are composed using weights $w_1$, $w_2$ and $w_3$.

%% file: sections/Algorithm.tex
\section{Proposed Algorithm}
\label{sec: Proposed Algorithm}
\subsection{Basic ideas}
\label{sec: Basic ideas}
 Let us consider solving an MMOP with $k$ equivalent Pareto subsets. Suppose that a weight vector $\boldsymbol{w}$ intersects with the Pareto front of the given MMOP at $\boldsymbol{p^*}$. As shown in Eq. (\ref{eq: MMOP explain}), we can find at most $k$ different Pareto optimal solutions $\{\boldsymbol{x}^*_1, \boldsymbol{x}^*_2\ldots\boldsymbol{x}^*_k\}$  in the decision space that are corresponding to $\boldsymbol{p^*}$. In addition, the scalarizing function values $g$ (with respect to $\boldsymbol{w}$) of these $k$ solutions are also equal to the minimum value (i.e., 0).

\begin{equation}
    \begin{gathered}
        \boldsymbol{F}(\boldsymbol{x}^*_1) = \boldsymbol{F}(\boldsymbol{x}^*_2) = \ldots = \boldsymbol{F}(\boldsymbol{x}^*_k) = \boldsymbol{p^*},\\
        g(\boldsymbol{w}, \boldsymbol{x}^*_1) = g(\boldsymbol{w}, \boldsymbol{x}^*_2) = \ldots = g(\boldsymbol{w}, \boldsymbol{x}^*_k)=0,\\
        \textbf{ s.t. } \forall \boldsymbol{x}^*_i, \boldsymbol{x}^*_j,  \boldsymbol{x}^*_i \neq \boldsymbol{x}^*_j  \text{ if } i \neq j.
    \end{gathered}
    \label{eq: MMOP explain}
\end{equation}

According to the above equations, for each weight vector $\boldsymbol{w}$, Eq. (\ref{eq: MMOP explain 2}) is a multi-modal single-objective optimization problem with $k$ global optimal solutions. In this way, the original MMOP is decomposed into a set of multi-modal single-objective optimization sub-problems. 

\begin{equation}
    \min_{\boldsymbol{x}} g(\boldsymbol{w}, \boldsymbol{x}).
    \label{eq: MMOP explain 2}
\end{equation}

Any multi-modal single-objective optimization algorithm can be used to optimize Eq. (\ref{eq: MMOP explain 2}). In our proposed algorithm, we apply a greedy iterative algorithm to optimize it. The detailed implementation of the proposed MMEA will be discussed in the next section.

Our proposed algorithm uses a similar sub-population framework to the above mentioned algorithms in Section \ref{sec: MOEA/D variants based on sub-population framework}. In our algorithm, each weight vector has a sub-population containing the same number of solutions. In real-world applications, it is challenging to specify the sub-population size in prior since usually the number of equivalent Pareto subsets is unknown. In Section \ref{sec: Influence of sub-population size}, we will further examine the performance of our proposed MMEA with different specifications of the sub-population size.
\subsection{Implementation}
\label{sec: Implementation}
Algorithm \ref{algorithm: Proposed MOEA/D-MM} shows the framework of our decomposition based MMEA called MOEA/D-MM (\underline{MOEA/D} for \underline{M}ulti-modal \underline{M}ulti-objective optimization). At the beginning, $\lambda = \floor{N / \mu }$ weight vectors are generated with the same method as MOEA/D. In line 3, $\mu$ solutions are randomly assigned to each weight vector. With this settings, each weight vector can preserve at most $\mu$ equivalent solutions. For convenience, solutions assigned to $\boldsymbol{w}_i$ are denoted as $P_i$.

\begin{algorithm}
\Parameter{$N$: population size; \\ $\mu$: sub-population size; \\ $g$: scalarizing function;}
\KwOut{Found solutions}
\tcc{The number of weight vectors}
$\lambda = \floor{N / \mu }$\;
Generate $\lambda$ weight vectors $\boldsymbol{W} = \{\boldsymbol{w}_1, \boldsymbol{w}_2, \ldots, \boldsymbol{w}_{\lambda}\}$\;
Randomly generate and assign $\mu$ solutions to each weight vector, i.e., $\boldsymbol{P} = \{P_1, \ldots, P_\lambda\}$ \;
Ideal point $\boldsymbol{z} = \{z_1, z_2, \ldots, z_M\}^T$\ where $z_i = \min_{x \in P}{f_i(\boldsymbol{x})}$\;
$T = \floor{\lambda/10}$\;
\Repeat{Termination criteria are met}{
    $\sigma = \textit{Estimate-Clearing-Radius}(\boldsymbol{P})$\;
    
    \ForEach{$\boldsymbol{w}_i \in \boldsymbol{W}$}{
        \tcc{Mating}
        $\boldsymbol{y} = \textit{Mating} (\boldsymbol{w}_i)$\; 
        \tcc{Update the ideal point $\boldsymbol{z}$}
        \ForEach{$j=1 \ldots M$}{
            $z_j \gets \min\{f_j(\boldsymbol{y}), z_j\}$\;
        }
        $S = P_i \cup \{\boldsymbol{y}\}$\;
        \tcc{Environmental selection}
        $P_i = \textit{Environmental-Selection}(\boldsymbol{w}_i, S, \sigma)$\; 
    }
}
$\boldsymbol{P}^{\prime} \gets$ non-dominated solutions in $\boldsymbol{P}$\;
\Return $\boldsymbol{P}^{\prime}$\;
\caption{Proposed MOEA/D-MM}
\label{algorithm: Proposed MOEA/D-MM}
\end{algorithm}

In each iteration, every sub-population is updated based on the $(\mu + 1)$ scenario. For a sub-population $P_i$, an offspring solution is generated by the procedures described in Algorithm \ref{algorithm: Mating}. Firstly, we randomly select a solution from $P_i$ as the first parent. To generate more diverse offspring solutions, the second parent is randomly selected from the union of neighborhood weight vectors' sub-populations. Then the generated offspring is added to $P_i$. In line 14, a solution is removed from $P_i$ based on the environmental selection procedure described in  Algorithm \ref{algorithm: Environmental-Selection}.

\begin{algorithm}
\Parameter{$\boldsymbol{w}_i$: input weight vector;}
\KwOut{Generated offspring;}
$\boldsymbol{W}^\prime \gets  T$ neighborhood weight vectors of $\boldsymbol{w}_i$\;
$B \gets$ the union of sub-populations of weight vectors in $\boldsymbol{W}^\prime$\;
$\boldsymbol{x}_1 \gets $ a randomly selected individual from $P_i$\;
$\boldsymbol{x}_2 \gets $ a randomly selected individual from $B$\;
\Return $\boldsymbol{y} \gets $ an offspring generated from $\{\boldsymbol{x}_1, \boldsymbol{x}_2\}$\;
\caption{\textit{Mating}}
\label{algorithm: Mating}
\end{algorithm}

In the environmental selection process, the clearing \cite{Petrowski1996clearing} method is introduced to create a niching structure in the decision space. The main idea of clearing is that within a given clearing radius $\sigma$, the best solution takes all resources, i.e., other solutions will be removed. In MOEA/D-MM, the clearing radius is estimated using Algorithm \ref{algorithm: Estimate-Clearing-Radius}. Before updating any sub-population, the clearing radius is set to the average Euclidean distance from each solution in the whole population to its $L$-th nearest neighbor in the decision space. In this paper, $L$ is set to $\floor{N/10}$.

Unlike the original clearing method, MOEA/D-MM only applies the clearing once in order to keep the sub-population size unchanged. In line 2 of Algorithm \ref{algorithm: Environmental-Selection}, a pair of points with the smallest distance between them in the current sub-population are found. If the Euclidean distance between them in the decision space is smaller than the clearing radius, the one with the better scalarizing function value will survive. Otherwise, the solution with the worst scalarizing function value in the current sub-population is removed.

\begin{algorithm}
\Parameter{$\boldsymbol{P}$: population;}
\KwOut{Clearing radius;}
\tcc{Neighborhood size}
$L = \floor{N/10}$\; 
$D \gets $ \{distance from $\boldsymbol{x}$ to its $L$-th nearest neighborhood solution in the decision space$|\boldsymbol{x}\in\boldsymbol{P}\}$\;
\Return $\sigma = \mean{D}$\;
\caption{\textit{Estimate-Clearing-Radius}}
\label{algorithm: Estimate-Clearing-Radius}
\end{algorithm}

\begin{algorithm}
\Parameter{$\boldsymbol{w}$: weight vector;\\$S$: candidate solutions;\\$\sigma$: clearing radius;}
\KwOut{Surviving solutions;}
    \tcc{scalarizing function value}
    $G = \{g(\boldsymbol{w}, \boldsymbol{x})|\boldsymbol{x}\in S\}$\;
    $\boldsymbol{x}_i, \boldsymbol{x}_j \gets$ the closest pair of points in $S$\;
    \uIf{$d(\boldsymbol{x}_i, \boldsymbol{x}_j) < \sigma$}{
        \tcc{Clearing in decision space}
        $\boldsymbol{x} \gets $ the solution with the worst $G$ value in $\{\boldsymbol{x}_i, \boldsymbol{x}_j\}$\;
    }
    \Else{
        \tcc{Greedy strategy}
        $\boldsymbol{x} \gets $ the solution with the worst $G$ value in $S$\;
    }
    $S \gets S\backslash{}\{\boldsymbol{x}\}$\;
    \Return $S$\;
\caption{\textit{Environmental-Selection}}
\label{algorithm: Environmental-Selection}
\end{algorithm}

\subsection{Effectiveness of MOEA/D-MM}
\begin{figure*}[t]
\centering
\begin{subfigure}[b]{.3\textwidth}
	\includegraphics[width=\linewidth]{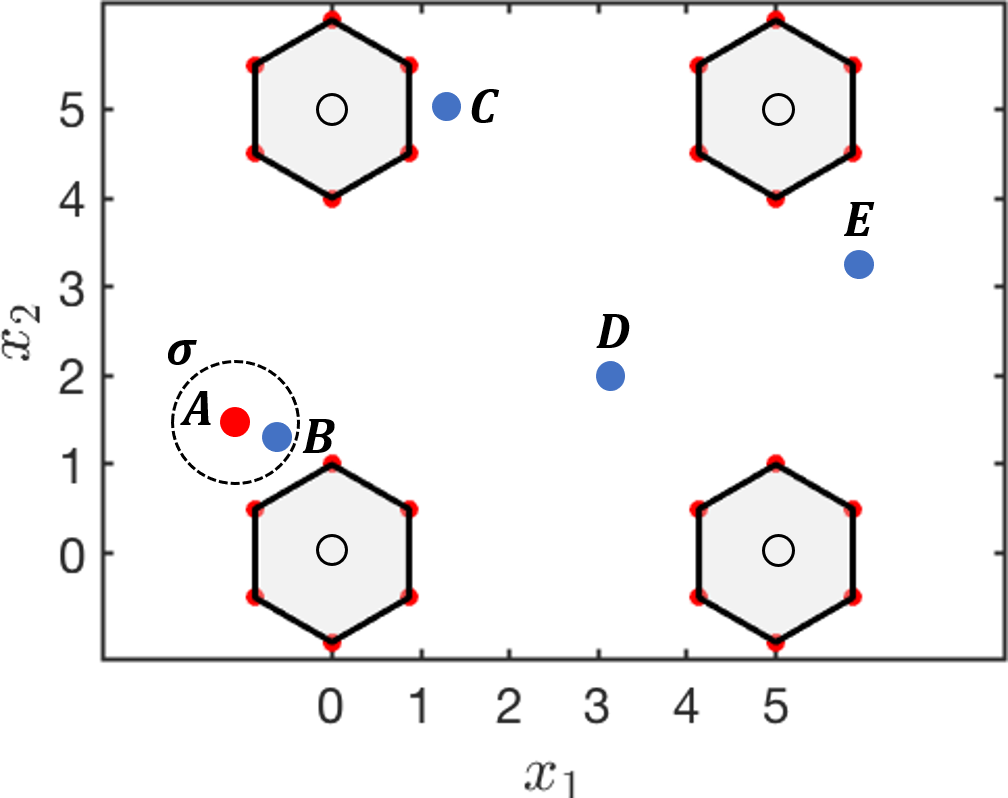}
	\caption{$A$ is removed.}
	\label{fig: Demo1}
\end{subfigure}
\hspace{0.05\textwidth}
\begin{subfigure}[b]{.3\textwidth}
	\includegraphics[width=\linewidth]{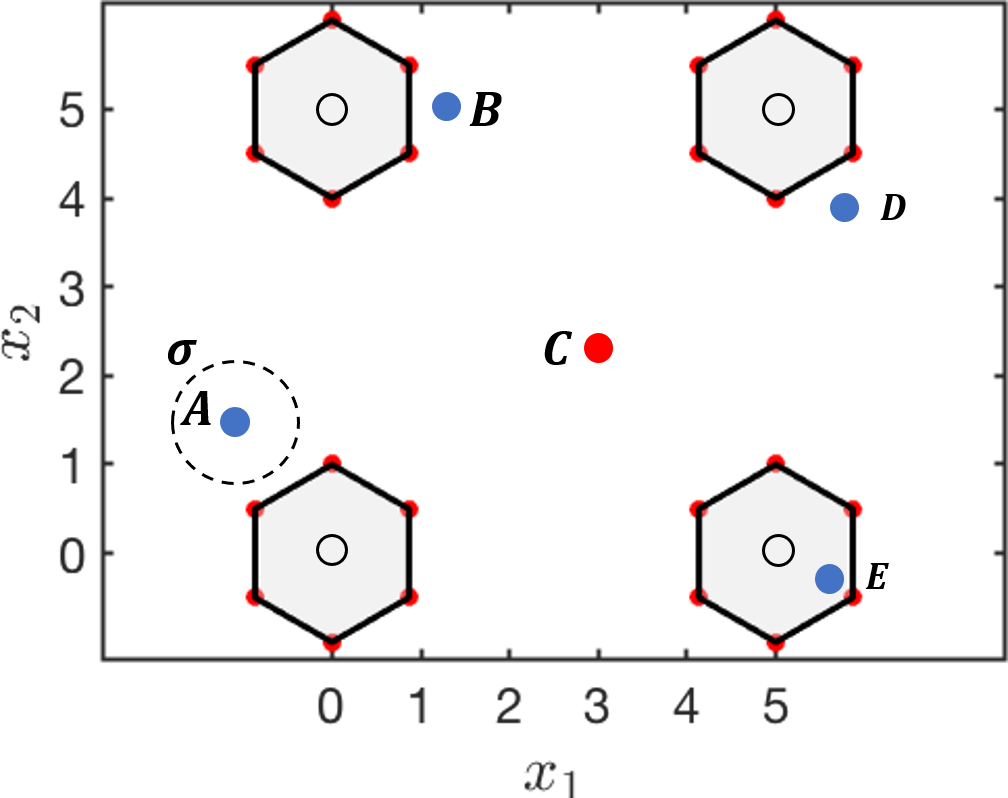}
	\caption{$C$ is removed.}
	\label{fig: Demo2}
\end{subfigure}
\par\bigskip
\begin{subfigure}[b]{.3\textwidth}
	\includegraphics[width=\linewidth]{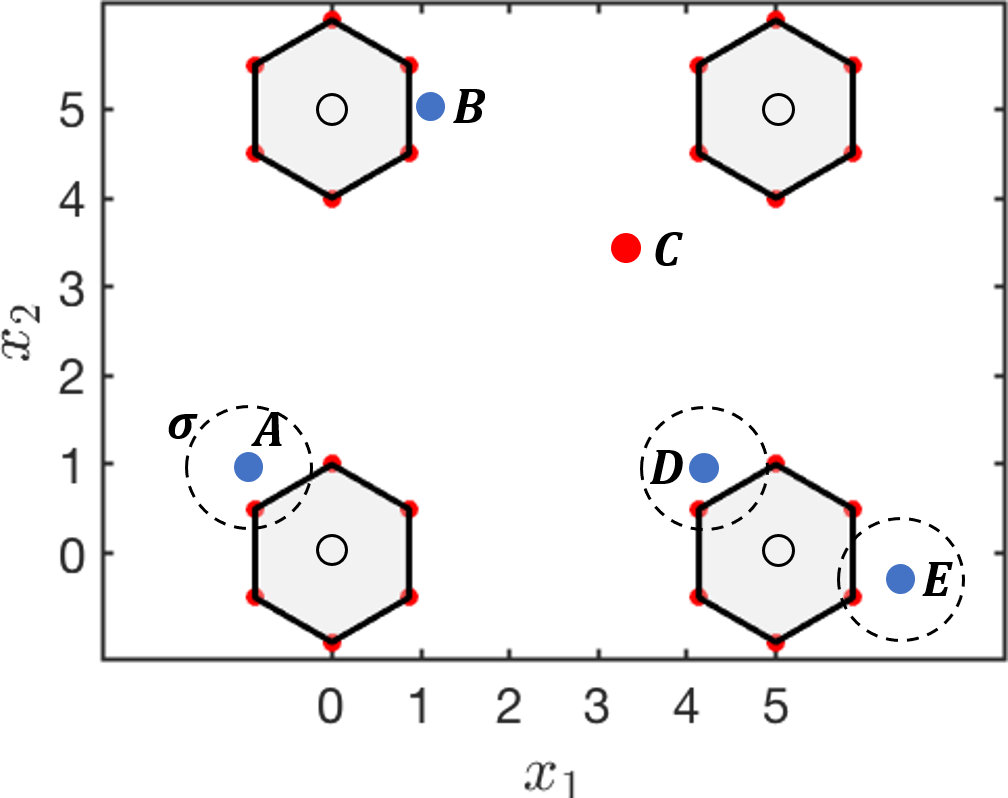}
	\caption{$C$ is removed.}
	\label{fig: Demo3}
\end{subfigure}
\hspace{0.05\textwidth}
\begin{subfigure}[b]{.3\textwidth}
	\includegraphics[width=\linewidth]{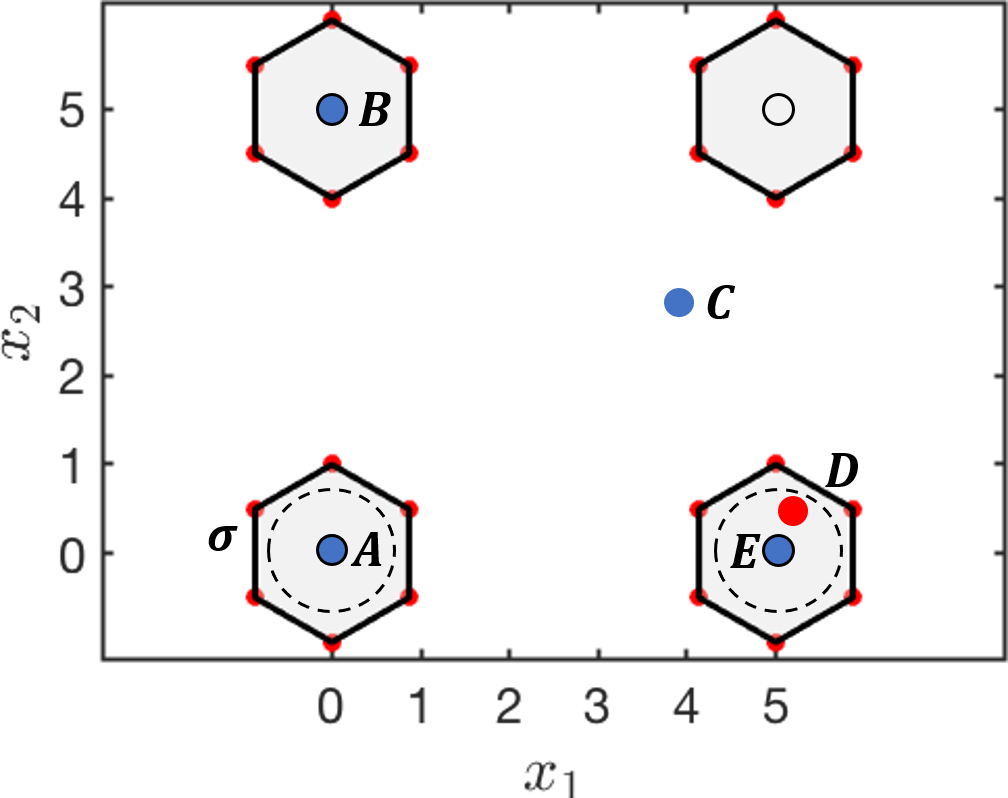}
	\caption{$D$ is removed.}
	\label{fig: Demo4}
\end{subfigure}
\caption{Illustration of the effect of the environmental selection mechanism of MOEA/D-MM on multi-polygon test problem.}
\label{fig: Effect of MOEA/D-MM}
\end{figure*}

In this section, we give some examples to illustrate the environmental selection mechanism in MOEA/D-MM. Fig. \ref{fig: Effect of MOEA/D-MM} shows a multi-polygon test problem\cite{ishibuchi2019salable} with four equivalent Pareto subsets. In this test problem, any solution inside the four hexagons (including solutions on the boundaries) is Pareto optimal. We assume that the sub-population size of each weight vector for MOEA/D-MM is $\mu = 4$. For clarity, we focus on the sub-population of the weight vector $\boldsymbol{w}$ whose scalarizing function value is minimized at the center of each hexagon. Therefore, solutions in this sub-population are searching toward the four equivalent Pareto optimal solutions located at the centers of the hexagons. In each generation, an offspring is generated and added to the sub-population of $\boldsymbol{w}$, and a solution in this sub-population is removed. In each figure in Fig. \ref{fig: Effect of MOEA/D-MM}, one of the five solutions (denoted by $(A, B, C, D, E)$) will be removed by the environmental selection mechanism. The dashed circle(s) in each figure represent the clearing radius. 

In Fig. \ref{fig: Effect of MOEA/D-MM} (\subref{fig: Demo1}), the distance between solutions $A$ and $B$ is smaller than the clearing radius $\sigma$. Therefore, $A$ is removed since it is worse than $B$. In some cases, the clearing does not remove any solution since all solutions in the current sub-population are not close to each other. Then the environmental selection is based on the greedy removal strategy, i.e., remove the worst solution in the current sub-population. For example, in Fig. \ref{fig: Effect of MOEA/D-MM} (\subref{fig: Demo2}), solution $C$ is removed. However, the greedy removal strategy cannot always make the best decision. In Fig. \ref{fig: Effect of MOEA/D-MM} (\subref{fig: Demo3}), solutions $D$ and $E$ are searching toward the same Pareto optimal solution in the right bottom hexagon. Although one of them is expected to be removed, the clearing mechanism does not work because the distance between solutions $D$ and $E$ is larger than $\sigma$. As opposed to our expectation, a potentially good solution $C$ is removed. Although the greedy strategy may make some mistakes, the clearing method can recover from the situation that several solutions in the same sub-population converge to the same Pareto optimal solution. After a number of iterations, the distance between solutions $D$ and $E$ will eventually become smaller than $\sigma$, then one of them will be removed. In Fig. \ref{fig: Effect of MOEA/D-MM} (\subref{fig: Demo4}), solution $D$ will be removed by the clearing mechanism. Notice that solution $C$ is Fig. \ref{fig: Effect of MOEA/D-MM} (\subref{fig: Demo4}) is a newly generated solution. With the clearing mechanism and the greedy removal strategy, MOEA/D-MM can efficiently preserve equivalent Pareto optimal solutions in sub-populations.

%% file: sections/ExperimentalResults.tex
\section{Experimental Studies}
\label{sec: Experimental Results}
\subsection{Experimental settings}
\subsubsection{Test problems}
The following four MMOPs are used to benchmark the performance of MMEAs: the SYM-PART problems\cite{Rudolph2007Capabilities}, the SSUF1 and SUF3 test problems\cite{liang2016multimodal} and the multi-polygon test problems\cite{ishibuchi2019salable}. Similar to \cite{peng2019multi}, we use the multi-polygon test problems to test the scalability of MOEA/D-MM regarding the dimension of the decision space. Parameters of the selected test problems are listed in Table \ref{table: Paramter settings}. In this table, $x_i$ denotes the $i$th decision variable, and the dimension of the objective space and the decision space of each test problem are denoted by $M$ and $D$, respectively. For the SYM-PART test problem, the length of each Pareto subset is $2a$, and the vertical and horizontal distances between the centers of two adjacent Pareto subsets are specified by $b$ and $c$, respectively.

\begin{table*}[t]
\caption{Parameter Settings For Selected Test Problems.}
\centering
\begin{tabular}{@{}cccccc@{}}
\toprule
& \multicolumn{5}{c}{Parameters} \\
\cmidrule(lr){2-6}
Problems & $M$ & $D$                    & Search Space                         & Special Parameters                                                                                      & Number of Pareto subsets \\ \midrule
SYM-PART          & 2 & 2                    & $x_i \in [-100, 100]^2$                & $a = 2, b=10, c=10$                                                                            & 9     \\
SSUF1             & 2 & 2                    & $x_1 \in [1, 3]$ and $x_2 \in [-1, 1]$ & -                                                                                           & 2     \\
SUF3              & 2 & 2                    & $x_1 \in [0, 1]$ and $x_2 \in [1, 2]$  & -                                                                                           & 2     \\
Multi-polygon    & 6 & $\{2, 4, 6, 8, 10\}$ & $x_i \in [-100, 100]^D$                & \begin{tabular}[c]{@{}c@{}}centers: ${(0, 0), (0, 5), (5, 0), (5, 5)}$\\ polygon radius = 1\end{tabular} & 4     \\ \bottomrule
\end{tabular}
\label{table: Paramter settings}
\end{table*}

\subsubsection{Performance indicators}
In our experiments, we use the modified inverted generational distance ($\operatorname{IGD}^+$) \cite{ishibuchi2015modified} and the IGDX\cite{zhou2009approximating} indicators for performance comparison. These two indicators are used to assess the quality of the obtained solution set in the objective and decision space, respectively. The $\operatorname{IGD}^+$ indicator improves the original IGD\cite{IGD} indicator by using a special distance function. In contrast to the original IGD indicator, the $\operatorname{IGD}^+$ indicator is weakly Pareto compliant. Formally, given a reference point set $P$ in the objective space, the $\operatorname{IGD}^+$ value of a set $A$ can be calculated using Eq. (\ref{eq: IGD+}).
\begin{equation}
    \begin{gathered}
        d^+(\boldsymbol{z}, \boldsymbol{a}) = \sqrt{\sum_{i=1}^m (\max\{a_i - z_i, 0\})^2},\\
        \operatorname{IGD^+}(A) = \frac{1}{|P|}\sum_{\boldsymbol{p}\in{}P}\min\{d^+(\boldsymbol{p}, \boldsymbol{a})|\forall \boldsymbol{a}\in A\}.
    \end{gathered}
    \label{eq: IGD+}
\end{equation}

The IGDX value of a set $A$ for a reference point set $S$ in the decision space is given by Eq. (\ref{eq: IGDX}).
\begin{equation}
    \operatorname{IGDX}(A) = \frac{1}{|S|}\sum_{\boldsymbol{x}\in{}S}\min\{d(\boldsymbol{x}, \boldsymbol{a})|\forall \boldsymbol{a}\in A\},
    \label{eq: IGDX}
\end{equation}
where $d$ is the Euclidean distance function.

For the IGDX indicator, the reference point set $S$ is generated by uniformly sampling 10,000 points on the Pareto sets of each test problem. Then, the image of $S$ in the objective space (i.e., $P$) is used to calculate the $\operatorname{IGD}^+$ indicator.

\subsubsection{Selected algorithms}
To verify the effectiveness of MOEA/D-MM on solving MMOPs, we compare its performance with the original MOEA/D with Tchebycheff and PBI scalarizing functions as well as three recently-proposed MMEAs: MOEA/D-AD\cite{tanabe2018decomposition}, DNEA\cite{liu2018double} and MO\_Ring\_PSO\_SCD\cite{yue2018multiobjective}. All algorithms are implemented on the PlatEMO\cite{PlatEMO} platform. Parameters for each algorithm are set to the suggested values in the original paper. In our experiments, each algorithm is evaluated with population size 300 and 100, 000 evaluations. All algorithms are examined on each test problem for 31 times in order to obtain reliable comparison results. For each algorithm on each test problem, a single run with the median HV over the 31 runs is selected for visualization.
\subsection{Experimental results}
\subsubsection{Comparison with MOEA/D}
\begin{figure}
    \centering
    \begin{subfigure}[b]{.24\textwidth}
        \includegraphics[width=\linewidth]{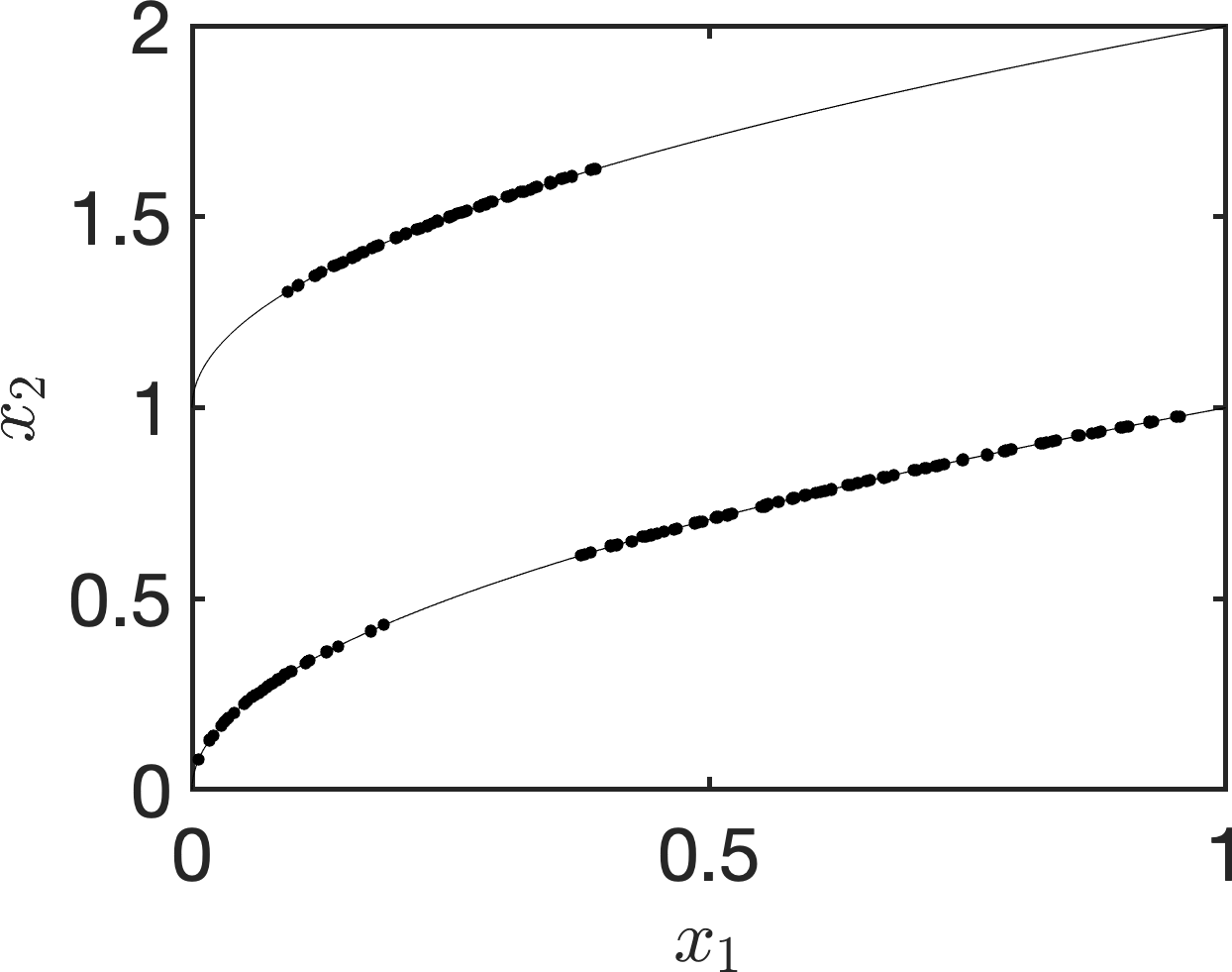}
	    \caption{MOEA/D}
    \end{subfigure}
    \begin{subfigure}[b]{.24\textwidth}
        \includegraphics[width=\linewidth]{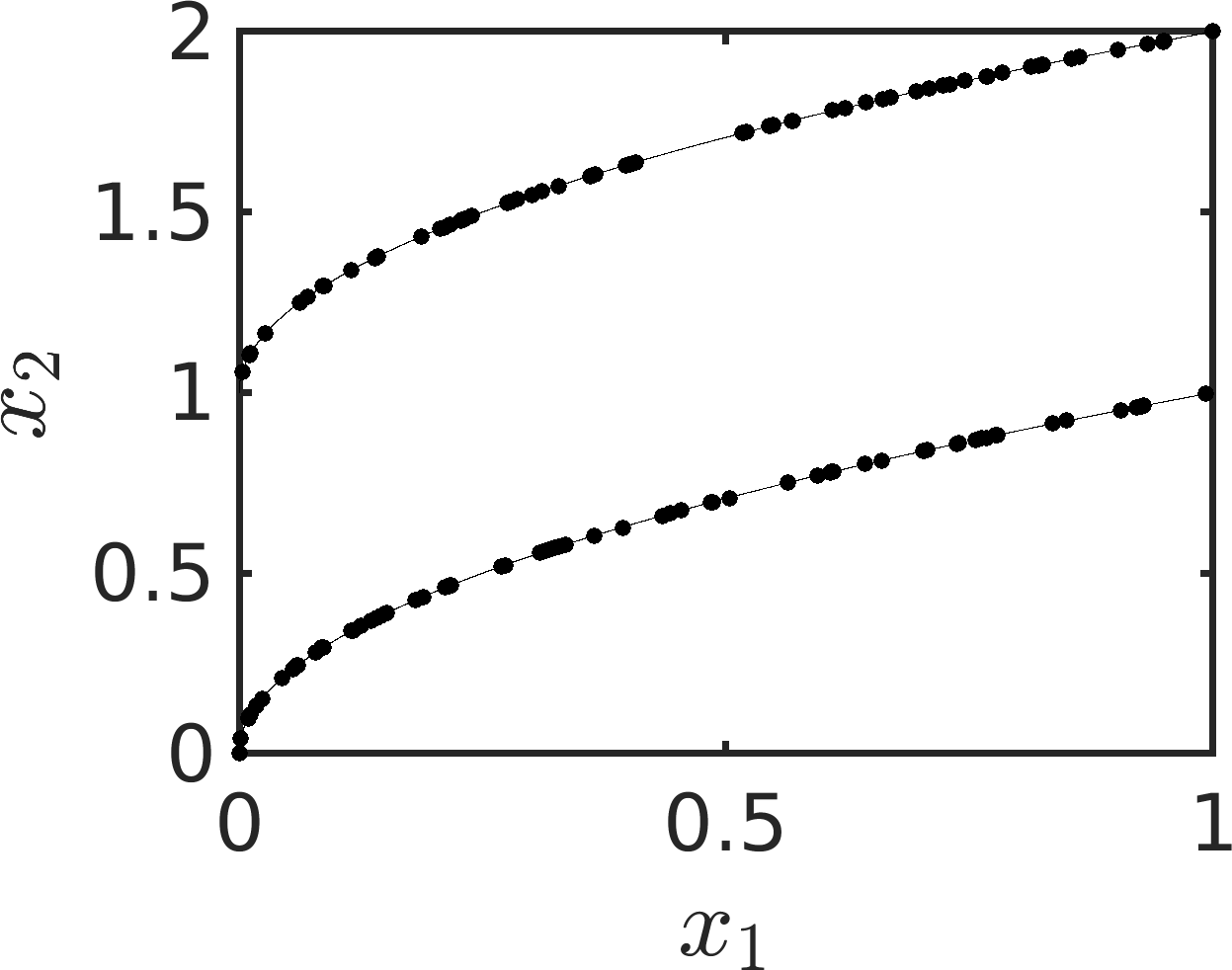}
	    \caption{MOEA/D-MM}
    \end{subfigure}
    \caption{Comparison of MOEA/D-MM and MOEA/D on the SUF3 test problem. In each figure, the thin lines denote the equivalent Pareto subsets, and the dark circles represent the obtained solutions.}
    \label{fig: comparison MOEA/D and MOEA/D-MM}
\end{figure}

In this section, we compare the performance of MOEA/D-MM and MOEA/D on the SUF3 test problem. Both of them use the Tchebycheff scalarizing function. In Fig. \ref{fig: comparison MOEA/D and MOEA/D-MM}, most solutions obtained by MOEA/D are located in one of the two Pareto subsets. However, MOEA/D-MM successfully locates all Pareto subsets. The simulation results clearly show that MOEA/D-MM can effectively preserve equivalent Pareto optimal solutions with the sub-population framework.
\subsubsection{Benchmark results on MMOPs}
In this section we specify the sub-population size as 4 in MOEA/D-MM. This specification is further discussed in the next section. 

\begin{figure*}
    \centering
    \begin{subfigure}[b]{.24\textwidth}
        \includegraphics[width=\linewidth]{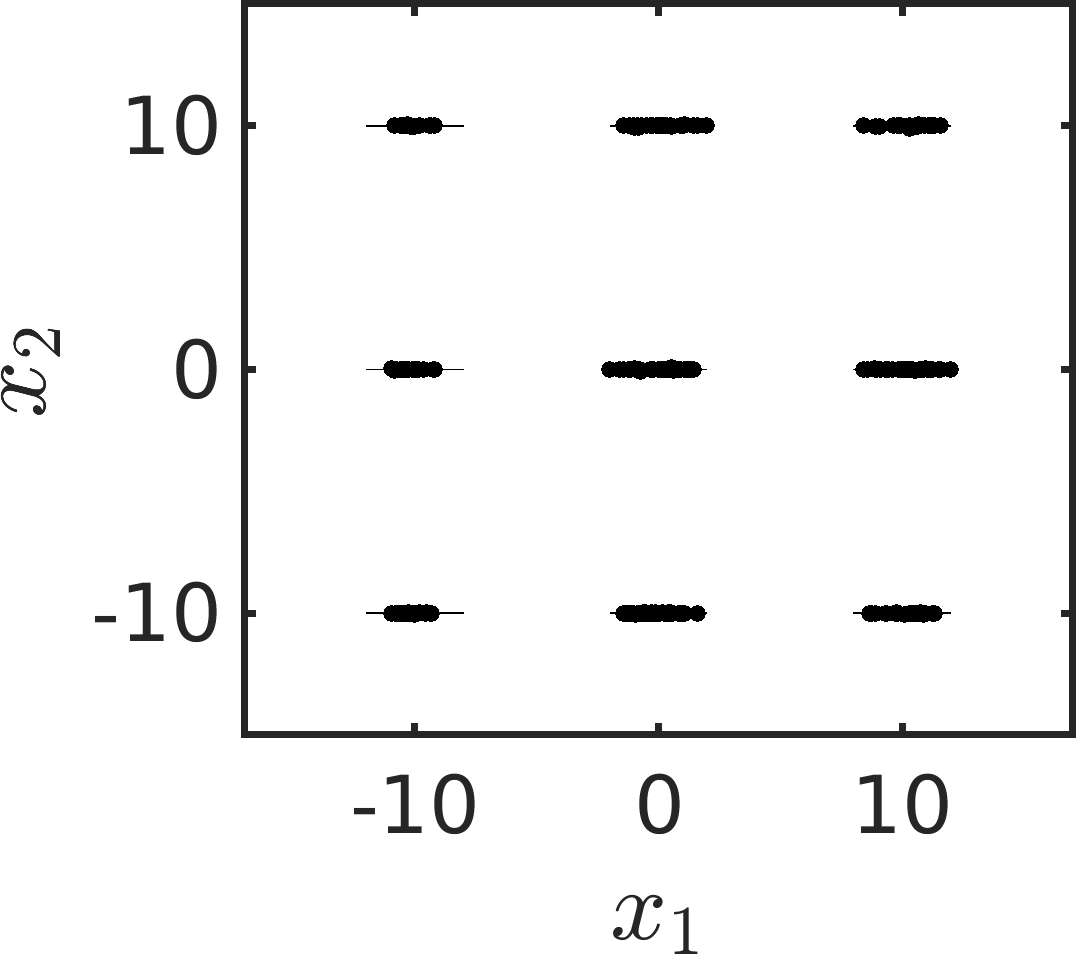}
	    \caption{SYM-PART.}
    \end{subfigure}
    \begin{subfigure}[b]{.24\textwidth}
        \raisebox{3mm}{\includegraphics[width=\linewidth]{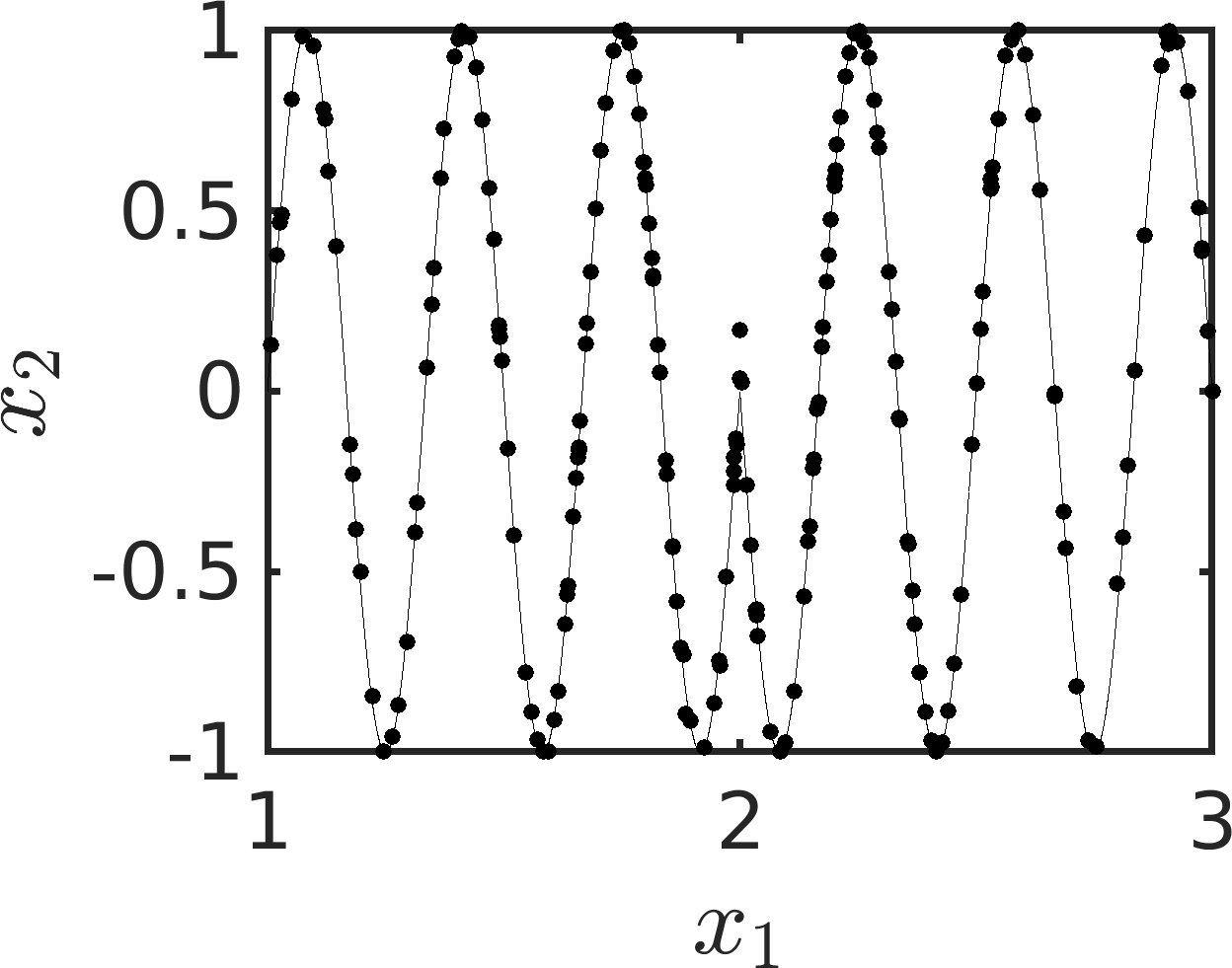}}
	    \caption{SSUF1.}
    \end{subfigure}
    \begin{subfigure}[b]{.24\textwidth}
        \raisebox{3mm}{\includegraphics[width=\linewidth]{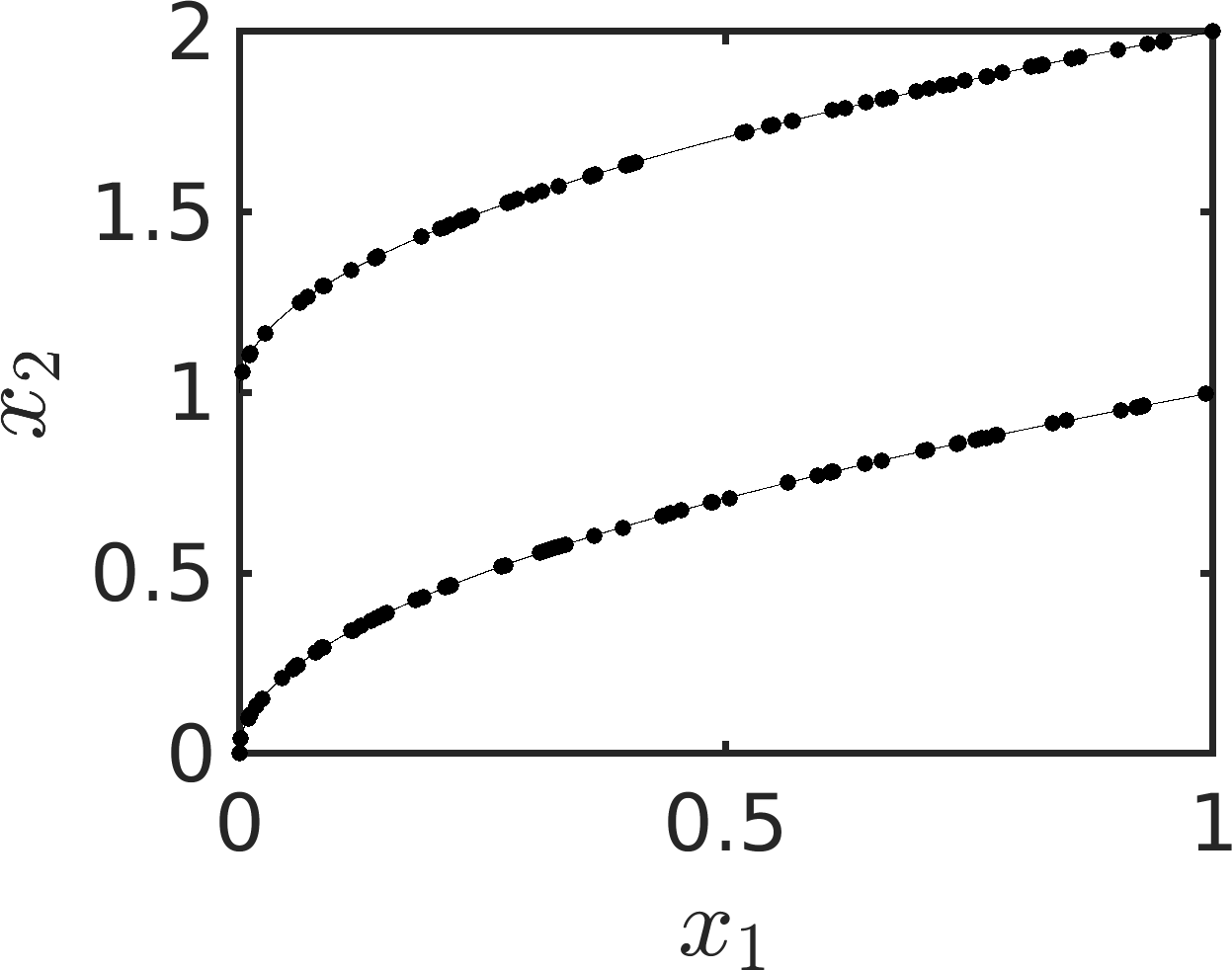}}
	    \caption{SUF3.}
    \end{subfigure}
    \begin{subfigure}[b]{.24\textwidth}
        \raisebox{3mm}{\includegraphics[width=\linewidth]{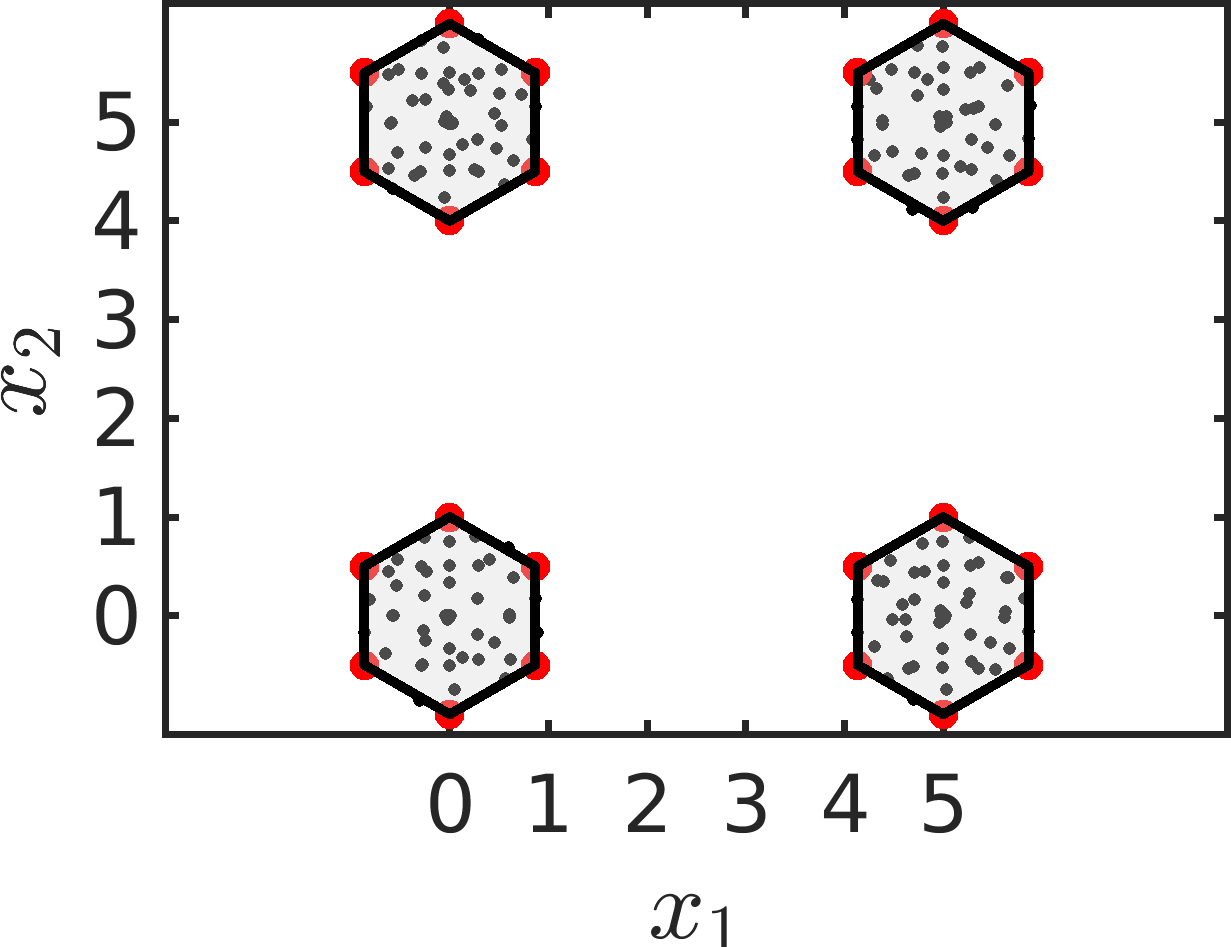}}
	    \caption{Multi-polygon($D=2$).}
    \end{subfigure}
    \caption{Visualization of non-dominated solutions in the final population of MOEA/D-MM with the Tchebycheff function in the decision space on each test problem.}
    \label{fig: MOEA/D-MM TCH visualization result}
\end{figure*}

\begin{figure*}
    \centering
    \begin{subfigure}[b]{.24\textwidth}
        \includegraphics[width=\linewidth]{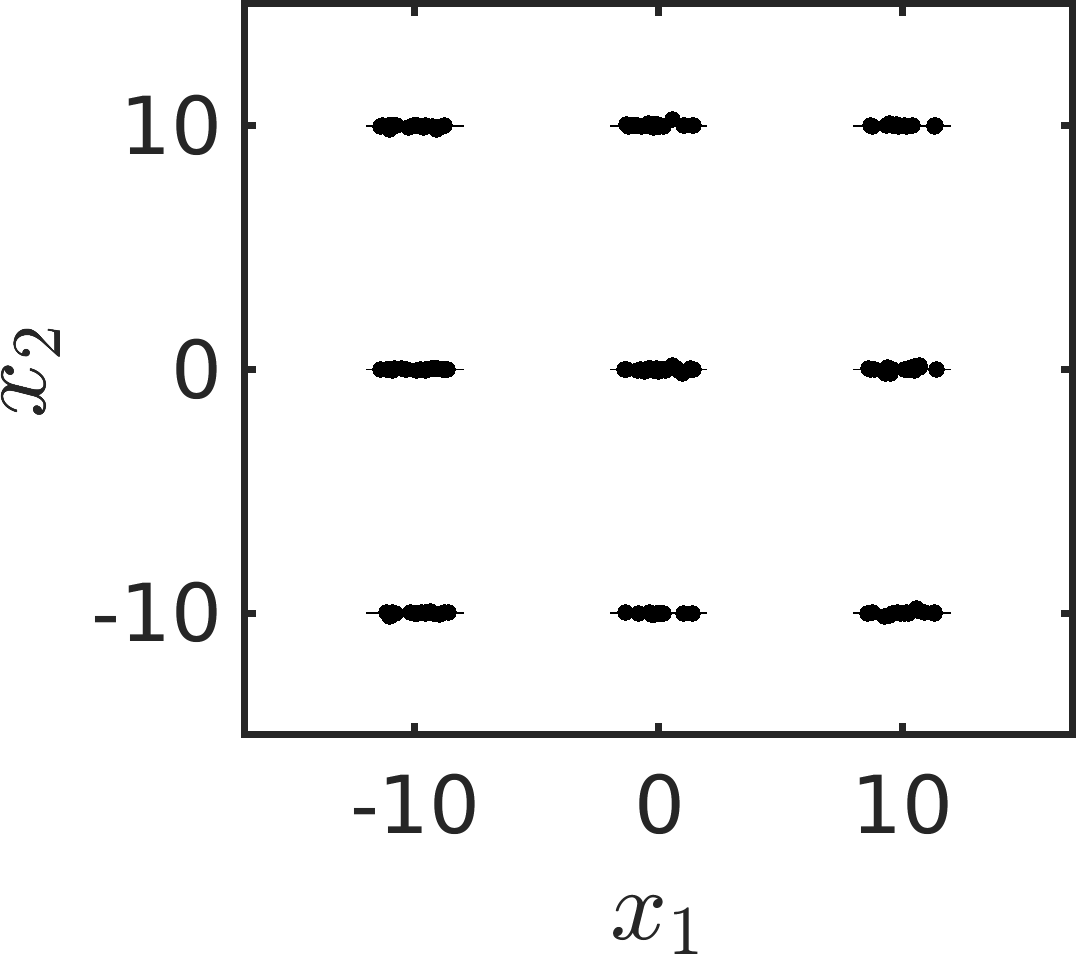}
	    \caption{SYM-PART.}
    \end{subfigure}
    \begin{subfigure}[b]{.24\textwidth}
        \raisebox{3mm}{\includegraphics[width=\linewidth]{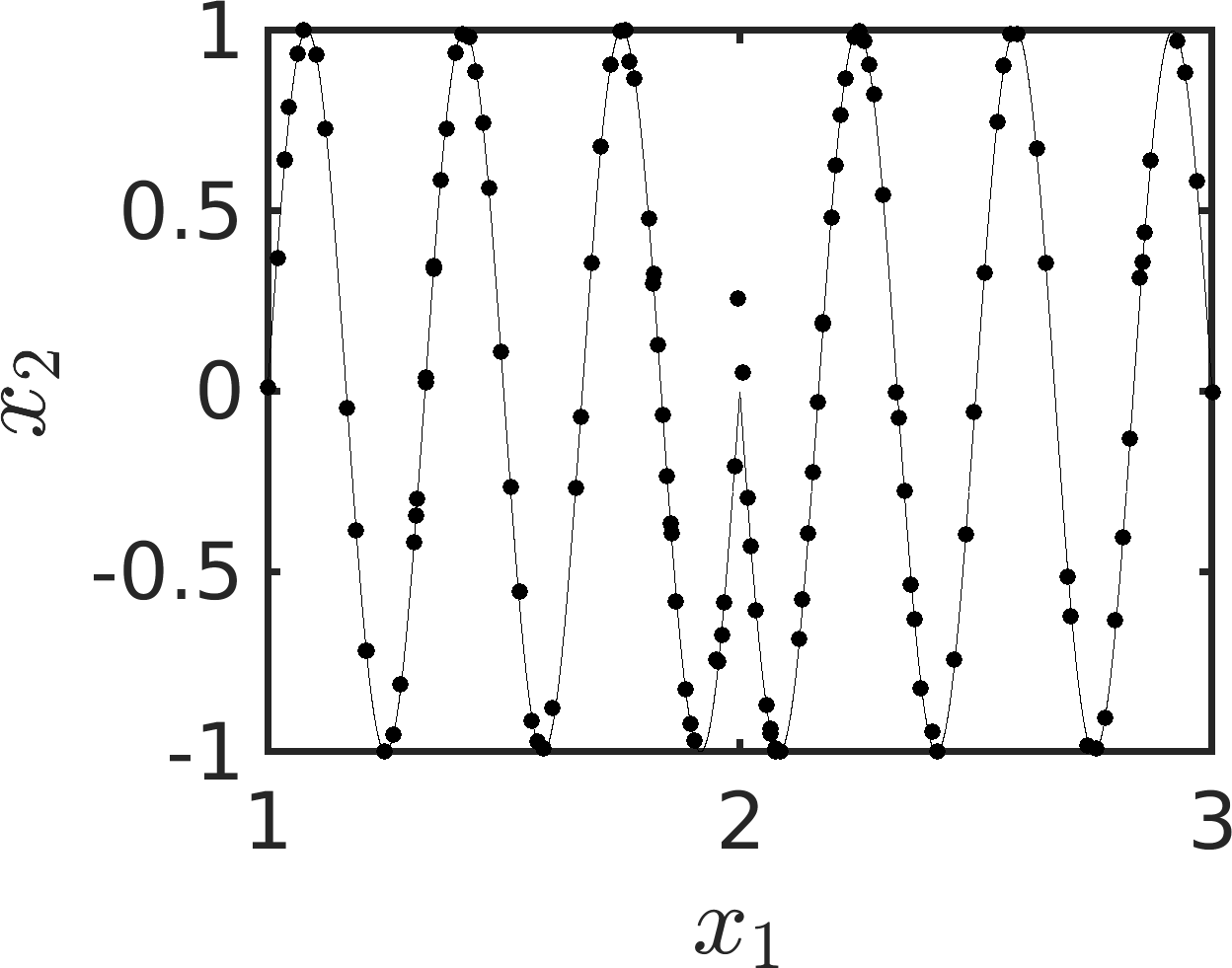}}
	    \caption{SSUF1.}
    \end{subfigure}
    \begin{subfigure}[b]{.24\textwidth}
        \raisebox{3mm}{\includegraphics[width=\linewidth]{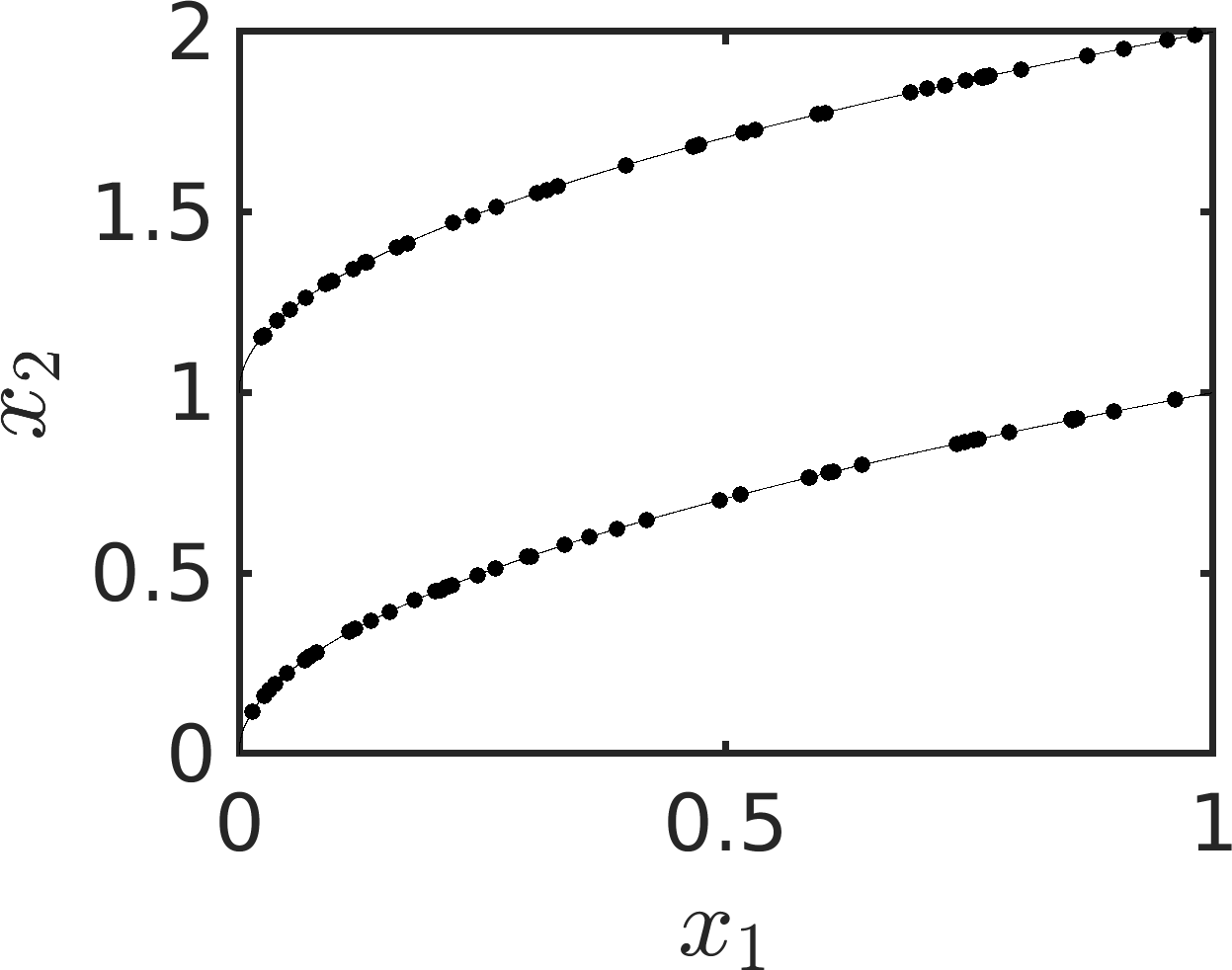}}
	    \caption{SUF3.}
    \end{subfigure}
    \begin{subfigure}[b]{.24\textwidth}
        \raisebox{3mm}{\includegraphics[width=\linewidth]{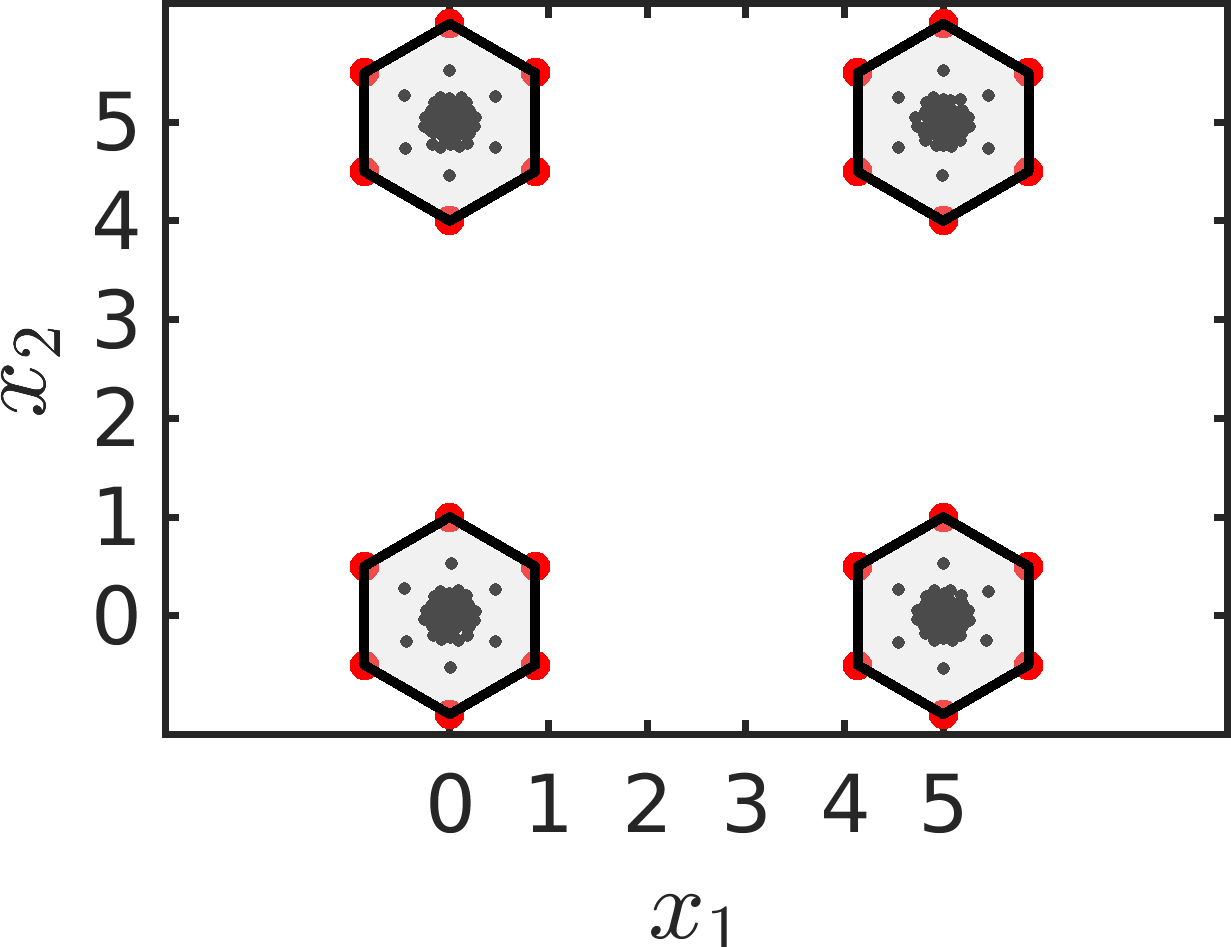}}
	    \caption{Multi-polygon($D=2$).}
	    \label{fig: MOEA/D-MM Polygons}
    \end{subfigure}
    \caption{Visualization of non-dominated solutions in the final population obtained by MOEA/D-MM with the PBI function in the decision space on each test problem.}
    \label{fig: MOEA/D-MM PBI visualization result}
\end{figure*}

Figs. \ref{fig: MOEA/D-MM TCH visualization result} and \ref{fig: MOEA/D-MM PBI visualization result} visualize the non-dominated solutions in the final population of MOEA/D-MM with Tchebycheff (MOEA/D-MM-TCH) and PBI (MOEA/D-MM-PBI) scalarizing functions in the decision space on each test problem, respectively. The black lines in Figs. \ref{fig: MOEA/D-MM TCH visualization result} (a)-(c) and Figs. \ref{fig: MOEA/D-MM PBI visualization result} (a)-(c) represent the Pareto set of the corresponding test problem. In Figs. \ref{fig: MOEA/D-MM TCH visualization result} (d) and \ref{fig: MOEA/D-MM PBI visualization result} (d), the four hexagons are equivalent Pareto subsets.  From the visualization results, we can see that MOEA/D-MM is able to obtain solution sets with very good coverage on all Pareto subsets.

Tables \ref{table: IGDX with MOEA/D-MM baseline} and \ref{table: IGD+ with MOEA/D-MM baseline} present the numerical comparison results among the seven algorithms. In each table, we use the Wilconxon rank-sum test with $p<0.05$ to compare each algorithm with MOEA/D-MM-TCH. The symbols $+$, $\approx$ and $-$ indicate that the corresponding algorithm is significantly better than, no significant difference from, and significantly worse than the performance of MOEA/D-MM-TCH, respectively. Best results in each table are highlighted.

As shown in Table \ref{table: IGDX with MOEA/D-MM baseline}, MOEA/D-MM-TCH has the best performance among the tested algorithms regarding the IGDX indicator, especially on the multi-polygon test problems with higher dimensional decision spaces, e.g. $D=4, 6, 8, 10$. Notice that MOEA/D-MM and MOEA/D with the PBI function perform worse than the Tchebycheff function almost on all test problems. Numerical results indicate that the Tchebycheff function can obtain more uniform solutions than the PBI function on selected test problems. The main reason is that these four test problems are all convex test problems. The PBI scalarizing function does not work well on such kind of problems. In particular, MO\_Ring\_PSO\_SCD performs the best on the SUF3, SSUF1 test problems and DNEA outperforms others on the SYM-PART test problem.

Table \ref{table: IGD+ with MOEA/D-MM baseline} shows the average $\operatorname{IGD}^+$ indicator value of each algorithm on each test problem. As observed in the literature\cite{liang2016multimodal}\cite{tanabe2018decomposition}, the IGD values of MMEAs are usually worse than MOEAs. This is because equivalent solutions are the same in the objective space, which means that they do not contribute to the IGD indicator. Since $\operatorname{IGD}^+$ is similar to IGD, MMEAs are expected to have worse $\operatorname{IGD}^+$ indicator values than MOEAs. Among all algorithms, MOEA/D with the Tchebycheff function achieves the best performance with respect to the $\operatorname{IGD}^+$ indicator. Among tested MMEAs, MO\_Ring\_PSO\_SCD and DNEA have the best performance on MMOPs with the 2-dimensional decision space while MOEA/D-MM-TCH outperforms other MMEAs with higher dimensional decision spaces in terms of the $\operatorname{IGD}^+$ indicator.

\begin{table*}[htbp]
\renewcommand{\arraystretch}{1.2}
\centering
\caption{Average IGDX values over 31 runs. Best results are highlighted.}
\begin{tabular}{cccccccccc}
\toprule
& & & \multicolumn{2}{c}{MOEA/D-MM} & & & & \multicolumn{2}{c}{MOEA/D}\\
\cmidrule(lr){4-5} \cmidrule(lr){9-10}
Problems&$M$&$D$&TCH&PBI&MO\_Ring\_PSO\_SCD&MOEA/D-AD&DNEA&TCH&PBI\\
\midrule
\multirow{1}{*}{SUF3}&2&2&1.2359e-2&2.0712e-2 $-$&\hl{7.3534e-3 $+$}&1.2853e-2 $\approx$&1.2541e-2 $\approx$&2.1853e-1 $-$&2.5902e-1 $-$\\
\hline
\multirow{1}{*}{SSUF1}&2&2&4.2628e-2&6.0374e-2 $-$&\hl{3.3075e-2 $+$}&4.8514e-2 $-$&3.5394e-2 $+$&1.3177e-1 $-$&1.4571e-1 $-$\\
\hline
\multirow{1}{*}{SYM-PART}&2&2&1.5503e-1&1.9154e-1 $-$&1.1212e-1 $+$&\hl{5.9621e-2 $+$}&5.7180e-1 $-$&1.8686e+0 $-$&9.7947e-1 $-$\\
\hline
\multirow{6}{*}{Multi-polygon}&6&2&1.0902e-1&2.1422e-1 $-$&1.1658e-1 $-$&1.7744e-1 $-$&\hl{1.0430e-1 $+$}&1.7188e-1 $-$&2.4330e-1 $-$\\
&6&4&\hl{2.4670e-1}&3.7535e-1 $-$&3.8229e-1 $-$&2.9541e-1 $-$&3.7152e+0 $-$&4.4101e+0 $-$&5.2624e+0 $-$\\
&6&6&\hl{4.5473e-1}&5.9612e-1 $-$&1.0294e+0 $-$&5.0676e-1 $-$&6.5989e+0 $-$&6.4156e+0 $-$&7.0966e+0 $-$\\
&6&8&\hl{7.3573e-1}&8.3894e-1 $-$&2.8267e+0 $-$&8.9567e-1 $-$&7.8238e+0 $-$&7.5401e+0 $-$&8.3309e+0 $-$\\
&6&10&1.1849e+0&\hl{1.1283e+0 $\approx$}&6.4177e+0 $-$&2.5597e+0 $-$&8.9284e+0 $-$&8.5160e+0 $-$&9.4045e+0 $-$\\
&6&100&3.0347e+1&3.2792e+1 $-$&2.9945e+2 $-$&8.0263e+1 $-$&\hl{2.9788e+1 $+$}&3.0377e+1 $-$&3.1208e+1 $\approx$\\
\hline
\multicolumn{3}{c}{$+/-/\approx$}&\textbf{Baseline}&0/8/1&3/6/0&1/7/1&3/5/1&0/9/0&0/8/1\\
\bottomrule
\end{tabular}
\label{table: IGDX with MOEA/D-MM baseline}
\end{table*}

\begin{table*}[htbp]
\renewcommand{\arraystretch}{1.2}
\centering
\caption{Average $\operatorname{IGD}^+$ values over 31 runs. Best results are highlighted.}
\begin{tabular}{cccccccccc}
\toprule
& & & \multicolumn{2}{c}{MOEA/D-MM} & & & & \multicolumn{2}{c}{MOEA/D}\\
\cmidrule(lr){4-5} \cmidrule(lr){9-10}
Problems&$M$&$D$&TCH&PBI&MO\_Ring\_PSO\_SCD&MOEA/D-AD&DNEA&TCH&PBI\\
\midrule
\multirow{1}{*}{SUF3}&2&2&4.1228e-3&6.5426e-3 $-$&2.8444e-3 $+$&5.0655e-3 $-$&\hl{1.8247e-3 $+$}&2.3882e-3 $+$&8.4060e-3 $\approx$\\
\hline
\multirow{1}{*}{SSUF1}&2&2&2.4583e-3&3.2304e-3 $-$&1.6090e-3 $+$&2.9579e-3 $-$&1.4054e-3 $+$&\hl{8.5769e-4 $+$}&1.0541e-3 $+$\\
\hline
\multirow{1}{*}{SYM-PART}&2&2&4.2798e-2&8.9186e-2 $-$&2.3064e-2 $+$&1.7412e-2 $+$&\hl{1.0714e-2 $+$}&1.3249e-2 $+$&7.3563e-2 $-$\\
\hline
\multirow{6}{*}{Multi-polygon}&6&2&9.0390e-2&1.8361e-1 $-$&6.6848e-2 $+$&1.6209e-1 $-$&\hl{5.0966e-2 $+$}&7.1261e-2 $+$&1.3933e-1 $-$\\
&6&4&1.3741e-1&2.6096e-1 $-$&1.9312e-1 $-$&2.2258e-1 $-$&1.4665e-1 $-$&\hl{1.1396e-1 $+$}&2.5126e-1 $-$\\
&6&6&2.2770e-1&3.9712e-1 $-$&6.0390e-1 $-$&3.1566e-1 $-$&3.0919e-1 $-$&\hl{1.6600e-1 $+$}&4.7236e-1 $-$\\
&6&8&3.5604e-1&5.7543e-1 $-$&1.6849e+0 $-$&4.8743e-1 $-$&4.7776e-1 $-$&\hl{2.3664e-1 $+$}&6.6033e-1 $-$\\
&6&10&5.0716e-1&7.2597e-1 $-$&4.1616e+0 $-$&7.1944e-1 $-$&6.3330e-1 $-$&\hl{3.0650e-1 $+$}&8.2931e-1 $-$\\
&6&100&1.7306e+1&1.4175e+1 $+$&7.0584e+2 $-$&1.6153e+2 $-$&8.1872e+0 $+$&\hl{3.5356e+0 $+$}&4.1586e+0 $+$\\
\hline
\multicolumn{3}{c}{$+/-/\approx$}&\textbf{Baseline}&1/8/0&4/5/0&1/8/0&5/4/0&9/0/0&2/6/1\\
\bottomrule
\end{tabular}
\label{table: IGD+ with MOEA/D-MM baseline}
\end{table*}
\subsubsection{Influence of sub-population size}
\label{sec: Influence of sub-population size}
In this section, we investigate the influence of the sub-population size on the performance of MOEA/D-MM. In MOEA/D-MM, the sub-population size determines the maximum number of equivalent solutions that a weight vector can preserve. Suppose we are handling an MMOP with $k$ equivalent Pareto subsets, and the sub-population size is $\mu$. Since the population size is $N$, the number of weight vectors is $\floor{N/\mu}$. When $\mu=k$, all equivalent solutions can be covered. When $\mu<k$, each weight vector can only preserve at most $\mu$ equivalent solutions. In this case, the decision maker has less choices when selecting a final solution. However, MOEA/D-MM may have a better search ability in the objective space since more weight vectors are used. When $\mu>k$, the performance of MOEA/D-MM will be deteriorated mainly due to the following two reasons. Firstly, the maximum number of equivalent solutions in each sub-population is not changed, which means that some solutions in the sub-population of each weight vector may have no contribution to the quality of the solution set. Secondly, a larger sub-population size means that less weight vectors are used, which will also reduce the search ability. 

\begin{figure}
    \centering
    \begin{subfigure}[b]{.24\textwidth}
        \includegraphics[width=\linewidth]{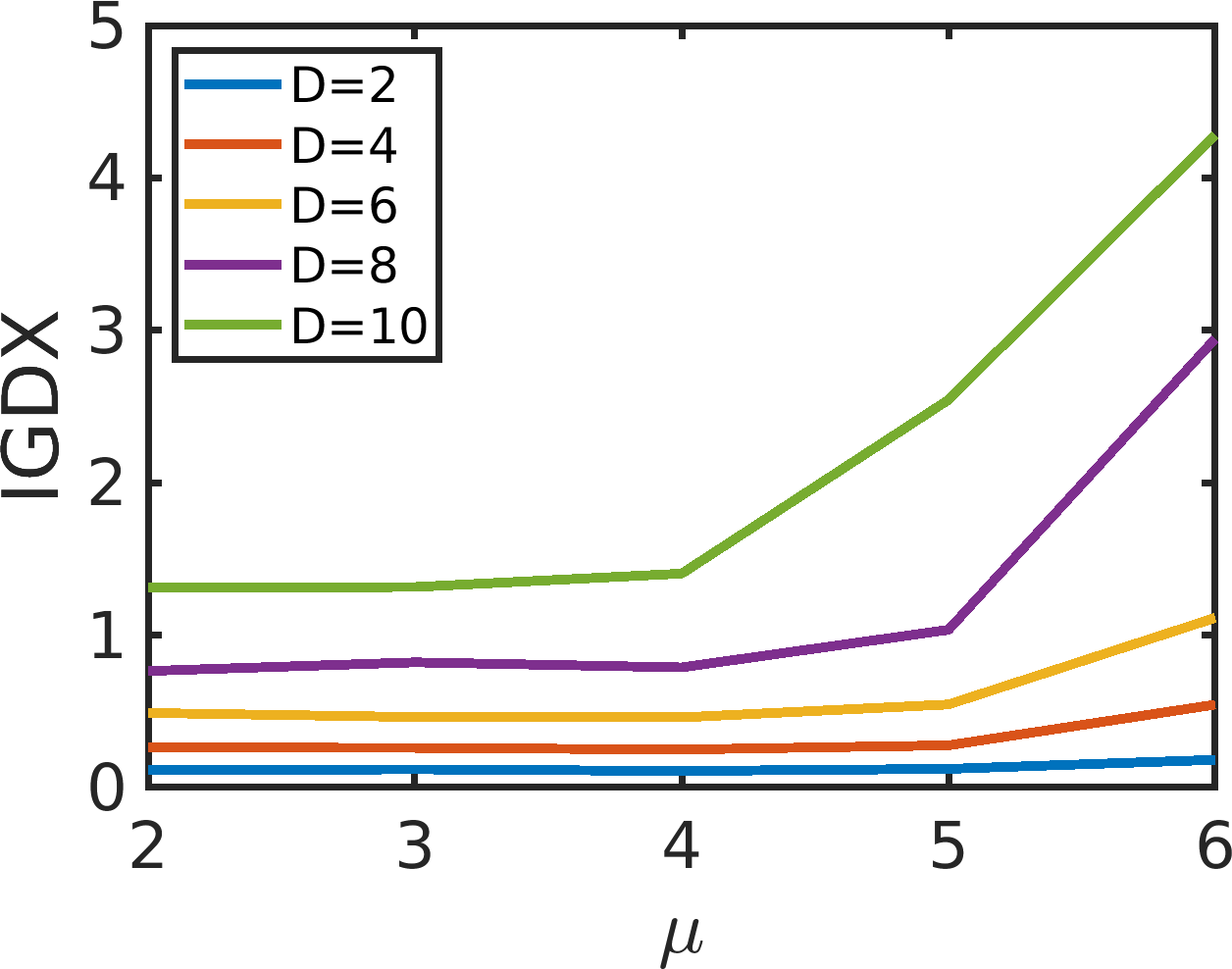}
        \caption{IGDX versus $\mu$.}
    \end{subfigure}
    \begin{subfigure}[b]{.24\textwidth}
        \includegraphics[width=\linewidth]{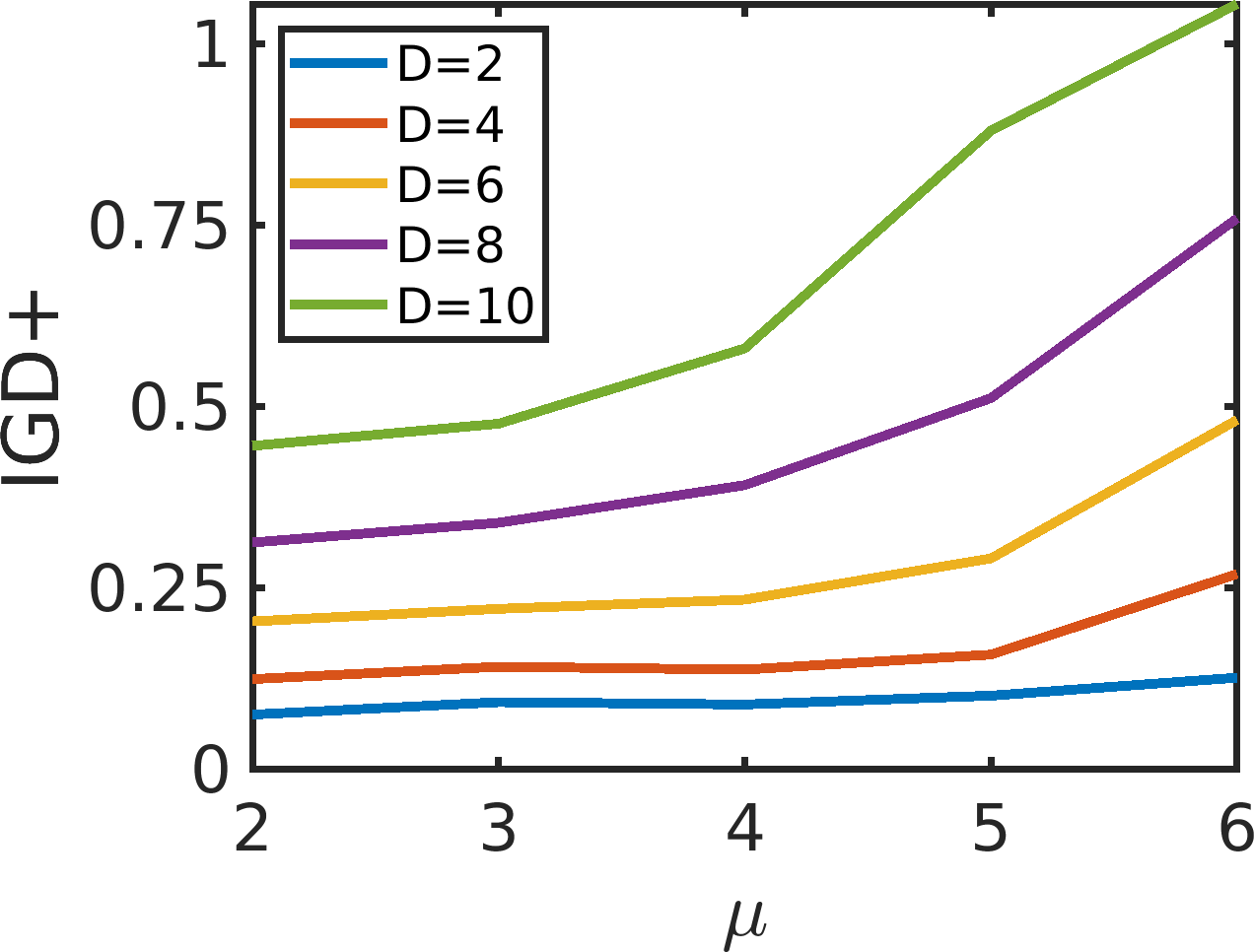}
         \caption{$\operatorname{IGD}^+$ versus $\mu$.}
    \end{subfigure}
    \caption{The average of IGDX and $\operatorname{IGD}^+$ indicator values with respect to sub-population size $\mu$.}
    \label{fig:change of IGDX and IGD+}
\end{figure}

\begin{figure}
    \centering
    \begin{subfigure}[b]{.24\textwidth}
        \includegraphics[width=\linewidth]{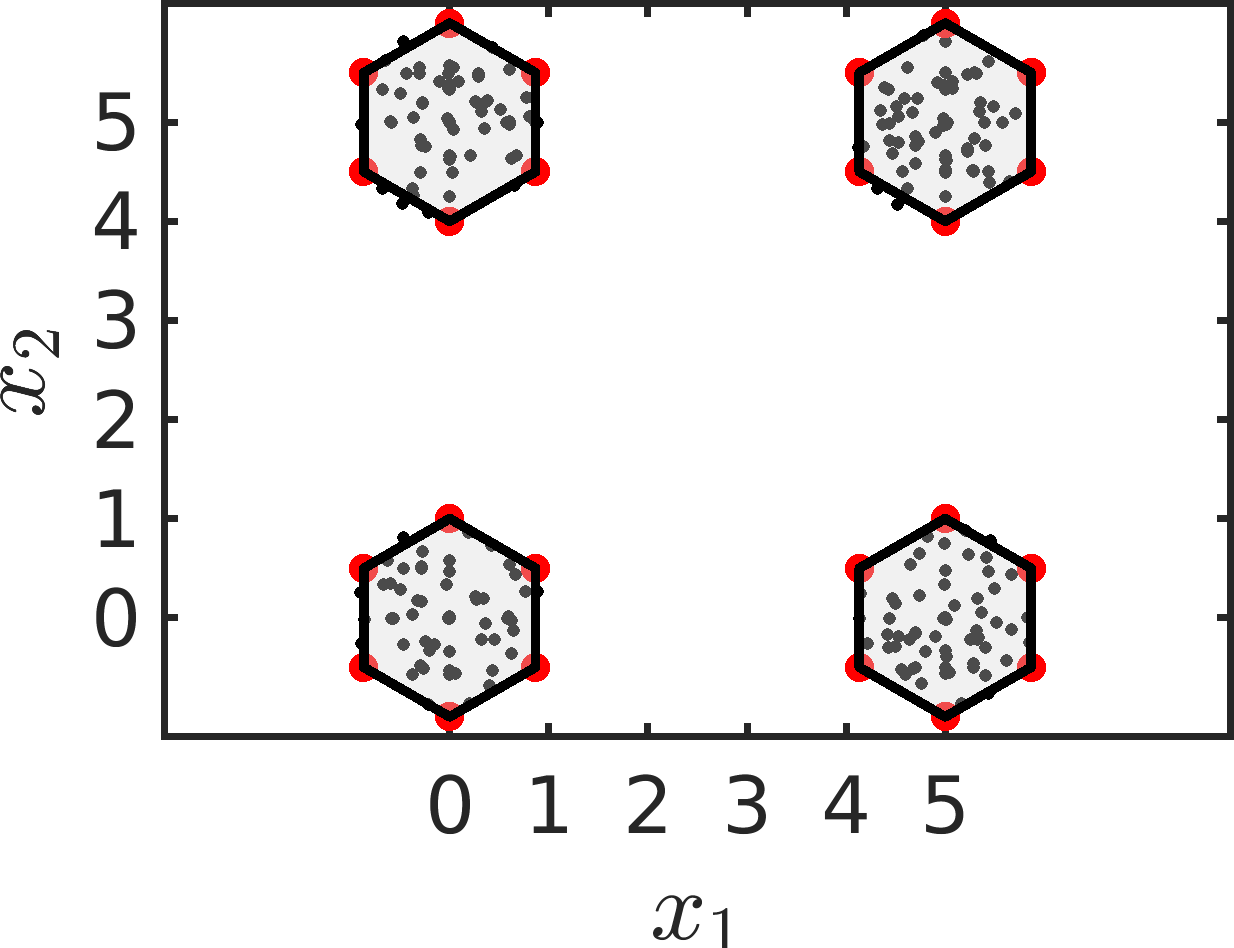}
        \caption{$\mu=2$}
    \end{subfigure}
    \begin{subfigure}[b]{.24\textwidth}
        \includegraphics[width=\linewidth]{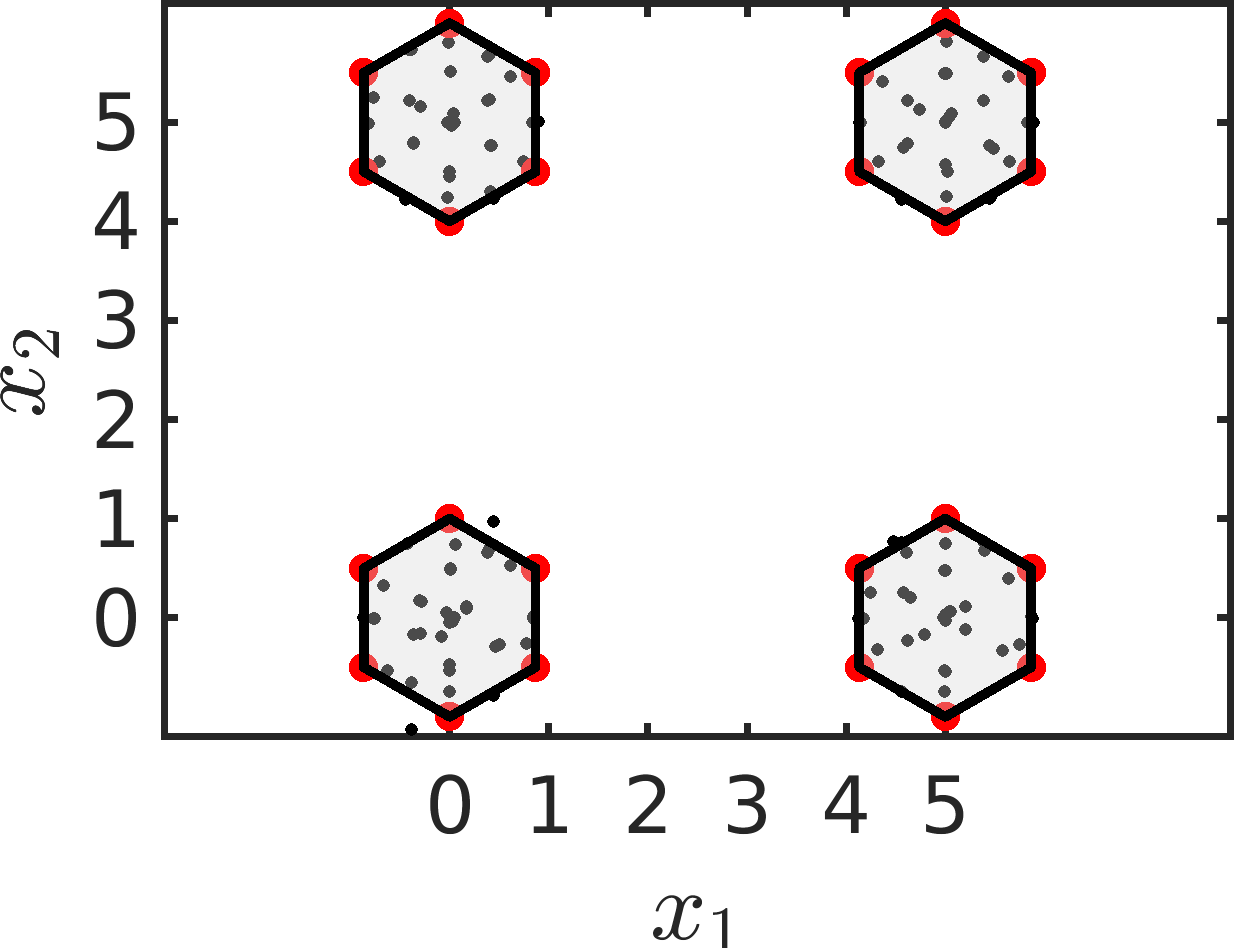}
        \caption{$\mu=6$}
    \end{subfigure}
    \caption{The final populations of MOEA/D-MM-TCH with two settings of $\mu$: $\mu=2$ and $\mu=6$.}
    \label{fig:visualization sub-population size}
\end{figure}

Fig. \ref{fig:change of IGDX and IGD+} shows the average IGDX and $\operatorname{IGD}^+$ indicator values obtained by MOEA/D-MM-TCH on the multi-polygon test problems with four equivalent Pareto subsets. Five specifications of $\mu$ ($\mu=2,3,4,5,6$) are examined for the multi-polygon test problems in five decision spaces ($D$-dimensional decision space for $D=2,4,6,8,10$). In Fig. \ref{fig:visualization sub-population size}, we show the final populations of MOEA/D-MM-TCH on the polygon test problem with two-dimensional decision space for two settings of $\mu$: $\mu=2$ and $\mu=6$. From Fig. \ref{fig:change of IGDX and IGD+} (a), when $\mu < 4$, IGDX values do not change too much. Because the IGDX indicator only measures the coverage of the Pareto set instead of the number of equivalent solutions. As shown in Fig. \ref{fig:visualization sub-population size} (a), when $\mu=2$, although a weight vector can only preserve two equivalent solutions, the coverage of Pareto subsets are almost the same as Fig. \ref{fig: MOEA/D-MM TCH visualization result} (d). This is because the number of weight vectors is two times as large as in the case of $\mu=4$. According to Fig. \ref{fig:change of IGDX and IGD+} (b), when $\mu>4$, the performance of the algorithm degrades in terms of the IGDX and $\operatorname{IGD}^+$ indicators. In Fig. \ref{fig:visualization sub-population size} (b), we can also observe that compared to $\mu=2$ and $\mu=4$, fewer non-dominated solutions are obtained when $\mu=6$ although all Pareto subsets are covered. The experimental results indicate that a smaller sub-population size is preferred in MOEA/D-MM.

%% file: sections/Conclusion.tex
\section{Concluding Remarks}
\label{sec: Concluding Remarks}
In this paper, we proposed MOEA/D-MM, a simple yet efficient multi-modal multi-objective optimization algorithm based on MOEA/D. We introduce a clearing mechanism and a greedy removal strategy to MOEA/D with the sub-population framework. The proposed algorithm shows promising performance in comparison with recently-proposed MMEAs on various MMOPs, especially on large-scale test problems. 

Several interesting research topics are left for future work. For example, a dynamic sub-population size adjustment strategy is needed for the handling of real-world optimization problems without knowledge about the number of Pareto sets. The development of new test problems and new performance indicators in the decision space is also an interesting research topic in the filed of multi-modal multi-objective optimization.